\RequirePackage[l2tabu, orthodox]{nag} 
\documentclass{article}

\usepackage[margin=1in]{geometry}
\usepackage{amsmath, amsthm, amssymb, graphicx}
\usepackage[dvipsnames]{xcolor}
\usepackage{url}

\DeclareMathOperator*{\TV}{TV}
\DeclareMathOperator*{\ar}{ar}
\DeclareMathOperator*{\vol}{vol}

\newcommand{\R}{\mathbb{R}}

\usepackage[normalem]{ulem}

\newtheorem{definition}{Definition}[section]

\newtheorem{theorem}[definition]{Theorem}

\usepackage{tikz}
\usetikzlibrary{svg.path}
\usepackage{pgfplots}
\usepgflibrary{fpu}
\usetikzlibrary{positioning,shadows,backgrounds}
\usepgfplotslibrary{fillbetween}

\definecolor{OIblack}{RGB}{0,0,0}
\definecolor{OIgreen}{RGB}{0,158,115}
\definecolor{OIblue}{RGB}{0,114,178}
\definecolor{OIbluelight}{RGB}{86,180,233}
\definecolor{OIyellow}{RGB}{240,228,66}
\definecolor{OIorange}{RGB}{230,159,0}
\definecolor{OIred}{RGB}{213,94,0}
\definecolor{OIpink}{RGB}{204,121,167}

\usepackage{wrapfig}

\title{Path Signatures for Feature Extraction \\ \large An Introduction to the Mathematics\\ Underpinning an Efficient Machine Learning Technique}

\author{Stephan Sturm\thanks{Department of Mathematical Sciences, Worcester Polytechnic Institute, 100 Institute Road, Worcester MA 06109, USA
	(e-mail: \texttt{ssturm@wpi.edu}).}
}

\begin{document}

\maketitle

\begin{abstract}
We provide an introduction to the topic of path signatures as means of feature extraction for machine learning from data streams. The article stresses the mathematical theory underlying the signature methodology, highlighting the conceptual character without plunging into the technical details of rigorous proofs. These notes are based on an introductory presentation given to students of the Research Experience for Undergraduates in Industrial Mathematics and Statistics at Worcester Polytechnic Institute in June 2024.
\end{abstract}

\vspace{5mm}
 
\begin{flushleft}
\end{flushleft}

\section{Introduction}

We are interested in the problems of classifying time series data into one or several pre-defined classes or using past data to predict future values. These are classical problems of supervised machine learning, and many methods have been developed to address it, e.g., logistic regression, support vector machines, decision trees, random forests and neural networks. Nothing about them will be said in this note, we assume they will be ready for further use. However, typically these methods cannot be applied to the time series directly, as the dimensionality of the data is much too high for that. One tries therefore in a first step to identify salient features of the time series relevant for the classification. This process is in the machine learning community known as \textit{feature extraction}. The choice of features could come from domain specific knowledge or be general statistical features such as volatility or moments as mean, variance, skewness and kurtosis. These work often reasonably well but miss some essential features of the time series. E.g., all general statistics mentioned do not take into account the order in which events happen. In particular for financial time series the order of events is usually quite important. For an easy example just think about the \textit{sequence risk} of an investor paying monthly into their retirement fund - for them it will matter greatly if the most successful years of the fund happen at the beginning of the investment period (and so affecting only a small portion of the investment) or at the end (where the capital paid during the entire duration can profit from them).

An alternative method of feature extraction is to consider the signature of (or a truncation thereof) paths generated from the time series. Signatures have been originally introduced by Kuo-Tsai Chen in the 1950s \cite{Che54, Che57} and gained importance and popularity through their role in Rough Path Theory and Rough Analysis developed by Terry Lyons in the 1990s \cite{L94, L98}. Recently (truncated) signatures have been used as tool for feature extraction in a variety of domains, such as character recognition, health care and finance \cite{Gra13, GLKF14, MKNSHL19}. Most literature on signatures is on a quite abstract level and needs an advanced mathematical understanding to tackle it. Practical implementation on the other hand are driven by the specifics of the problem studied and often use signatures just as \textit{black box}. The goal of this note is to give a gentle introduction to signatures as tool for feature extraction for readers that have as background only basic real analysis or advanced calculus. While mathematically not completely rigorous, we aim to highlight a conceptual understanding of the method that provides the necessary background for successful practical applications of the signature approach. While our presentation is guided by the problem of time series classification, signatures can be used in the same way as feature set for prediction problems.

\section{Motivation: Taylor Series}

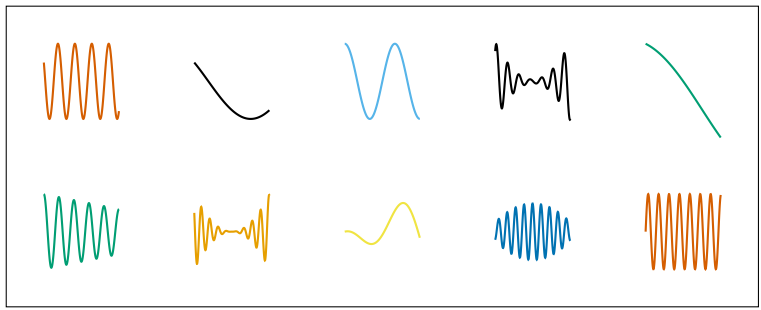
\begin{wrapfigure}{R}{0.65\textwidth}\centering
	\begin{tikzpicture}
		\draw[black] (0,-2) -- (0,2) -- (10,2) -- (10,-2) -- cycle;
		\begin{scope}[xshift = 0.5cm, yshift = 1cm, scale =0.5]
				\draw[OIred, thick] plot[domain=0:2,samples=400] (\x,{(sin(800*(\x)+150))});
		\end{scope}
		\begin{scope}[xshift = 0.5cm, yshift = -1cm, scale =0.5]
				\draw[OIgreen, thick] plot[domain=0:2,samples=400] (\x,{(cos(700*(\x)+200*\x))*(1-0.2*\x)});
		\end{scope}
		\begin{scope}[xshift = 2.5cm, yshift = 1cm, scale =0.5]
				\draw[OIblack, thick] plot[domain=0:2,samples=400] (\x,{(sin(80*(\x)+150))});
		\end{scope}
		\begin{scope}[xshift = 2.5cm, yshift = -1cm, scale =0.5]
				\draw[OIorange, thick] plot[domain=0:2,samples=400] (\x,{(sin(1600*(\x)+150))*(1-\x)^2});
		\end{scope}
		\begin{scope}[xshift = 4.5cm, yshift = 1cm, scale =0.5]
				\draw[OIbluelight, thick] plot[domain=0:2,samples=400] (\x,{(cos(270*(\x)))});
		\end{scope}
		\begin{scope}[xshift = 4.5cm, yshift = -1cm, scale =0.5]
				\draw[OIyellow, thick] plot[domain=0:2,samples=400] (\x,{(sin(200*(\x)+150))*(0.5*(\x)});
		\end{scope}
		\begin{scope}[xshift = 6.5cm, yshift = 1cm, scale =0.5]
				\draw[OIblack, thick] plot[domain=0:2,samples=400] (\x,{(sin(1200*(\x)+50))*(0.3+(0.5-\x)*(1.5-\x)});
		\end{scope}
		\begin{scope}[xshift = 6.5cm, yshift = -1cm, scale =0.5]
				\draw[OIblue, thick] plot[domain=0:2,samples=400] (\x,{(sin(800*(\x)+150))*(cos(800*(\x)+150))*(0.5+(\x)*(2-\x))});
		\end{scope}
		\begin{scope}[xshift = 8.5cm, yshift = 1cm, scale =0.5]
				\draw[OIgreen, thick] plot[domain=0:2,samples=400] (\x,{(cos(60*(\x)))-0.5*\x});
		\end{scope}
		\begin{scope}[xshift = 8.5cm, yshift = -1cm, scale =0.5]
				\draw[OIred, thick] plot[domain=0:2,samples=400] (\x,{(sin(1300*(\x)+0.75*\x))});
		\end{scope}
		\end{tikzpicture}
	\caption{How can we train a computer that it can distinguish between wiggly and smooth curves?}\label{fig:mot}
\end{wrapfigure}

To understand the ideas behind the signature approach, we look first at a completely different problem. Assume you have the task to differentiate two classes of curves, some only slightly bent and some quite wiggly. How should you approach such a problem? Humans can do this task easily and intuitively understand how to do the classification. But it is not so straightforward to teach this process to a computer. Automated decision making will have to take a different direction. One possible approach is using Taylor series: The Taylor coefficients describe the characteristics at the curve locally around a point: location, slope, convexity, etc. In their combination these can work as features for a classification algorithm. 

While many readers might be familiar with Taylor series, we will nevertheless develop them from scratch as this will be insightful for the future development of signature. As starting point we consider the Fundamental Theorem of Calculus: For a continuous differentiable function $f: [a,b] \to \R$ we have 
\begin{equation}\label{eq:A0}
f(x) = f(x_0) + \int_{x_0}^x f'(y) \, dy
\end{equation}
for $x, x_0 \in [a,b]$. This by itself can already be understood as first (actually \textit{zeroth order}) approximation of the function $f$ at the point $x$: as long  as $x$ is close to $x_0$, $f(x_0)$ is a reasonable approximation for $f(x)$, $f(x) \approx f(x_0)$, and the error we are making is described by the remainder term, $\mathcal{R}_0(x;x_0) = \int_{x_0}^x f'(y) \, dy$.

To get a better approximation, we can use the fundamental theorem of calculus again, however now applied to the function $f'$ (which we assume to be continuously differentiable on $[a,b]$):
\[
f'(x) = f'(x_0) + \int_{x_0}^x f''(y) \, dy.
\]
Plugging this into equation \eqref{eq:A0} we get
\begin{align}\label{eq:A1}
f(x) & = f(x_0) + \int_{x_0}^x \biggl(f'(x_0) + \int_{x_0}^y f''(z) \, dz \biggr)\, dy \nonumber\\
&= f(x_0) + f'(x_0)\int_{x_0}^x 1 \, dy + \int_{x_0}^x \int_{x_0}^y f''(z) \, dz dy\\
&= f(x_0) + f'(x_0)\bigl(x-x_0\bigr) + \int_{x_0}^x \int_{x_0}^y f''(z) \, dz dy \nonumber
\end{align}
Here we have a linear or \textit{first order} approximation $f(x) \approx f(x_0) + f'(x_0)\bigl(x-x_0\bigr)$: the value of $f$ at $x$ equals approximately the value of a linear function at $x$, a function that goes through the point $\bigl(x_0, f(x_0\bigr)$ and has the same slope as $f$ itself at the point $x_0$, namely $f'(x_0)$. The error of this approximation is here given by the remainder term $\mathcal{R}_1(x;x_0) = \int_{x_0}^x \int_{x_0}^y f''(z) \, dz dy$. While this is the typical interpretation, for our purposes it makes more sense to focus on the form before the final simplification, namely \eqref{eq:A1}. 

We can continue this approach: given that $f$ is three times continuously differentiable on $[a,b]$ we have
\[
f''(x) = f''(x_0) + \int_{x_0}^x f'''(y) \, dy
\]
and plugging this into equation $\eqref{eq:A1}$ yields the second order approximation
\begin{align}\label{eq:A2}
f(x) & =  f(x_0) + f'(x_0)\int_{x_0}^x 1 \, dy + f''(x_0) \int_{x_0}^x \int_{x_0}^y 1 \, dz dy + \int_{x_0}^x \int_{x_0}^y \int_{x_0}^z f'''(w) \, dw dz dy.
\end{align}
The error of the approximation 
\[
f(x)  \approx  f(x_0) + f'(x_0)\int_{x_0}^x 1 \, dy + f''(x_0) \int_{x_0}^x \int_{x_0}^y 1 \, dz dy =  f(x_0) + f'(x_0)\bigl(x-x_0\bigr) + f''(x_0)\frac{(x-x_0)^2}{2} 
\]
is here given by the remainder term $\mathcal{R}_2(x;x_0) = \int_{x_0}^x \int_{x_0}^y \int_{x_0}^z f'''(w) \, dw dz dy$. Continuing the expansion ad infinitum we can expect
\begin{align}\label{eq:Ainf}
f(x) & =  \sum_{k =0}^\infty f^{(k)}(x_0) \int_{x_0}^x \int_{x_0}^{x_k} \cdots \int_{x_0}^{x_3} \int_{x_0}^{x_2} 1 \, dx_1 dx_2 \cdots  dx_{k-1} dx_k\\
&=  \sum_{k =0}^\infty f^{(k)}(x_0) \frac{(x-x_0)^k}{k!} \nonumber
\end{align}
as long as 
\[
\lim_{n \to \infty} \mathcal{R}_n(x;x_0) = \lim_{n \to \infty} \int_{x_0}^x \int_{x_0}^{x_{n+1}} \int_{x_0}^{x_n} \cdots \int_{x_0}^{x_3} \int_{x_0}^{x_2} f^{(n+1)}(x_1) \, dx_1 dx_2 \cdots dx_{n-1} dx_n dx_{n+1}= 0.
\]
Indeed, the representation of the iterated integral as monomial can be proved by induction. We see that the function $f$ is characterized fully by the \textit{Taylor coefficients}
\[
a_n = \frac{f^{(n)}(x_0)}{n!},
\]
so knowing the function $f$ is equivalent to knowing the sequence
\[
\bigl(a_0, a_1, a_2, \cdots\bigr).
\]
Realistically, we cannot use the whole sequence as feature but will have to truncate it at some point, using 
\[
\bigl(a_0, a_1, a_2, \cdots, a_{N-1}, a_N\bigr).
\]
as feature to be used by the classification algorithms. Practically this will work well for differences in "wiggliness" around the base point $x_0$. If we do not trust that local information around a single point is good enough, we can just use averages over different points along the curve as feature on which we base the classification.

\begin{wrapfigure}{R}{0.42\textwidth}\centering
	\begin{tikzpicture}[scale=2]
		\draw[gray!80, ->] (-1.7cm,0cm) -- (1.7cm,0cm) node at (1.8,-0.1) {\small $x$};  
		\draw[gray!80, ->] (0cm,-0.2cm) -- (0cm,0.6cm) node at (-0.1,0.55) {\small $y$};  
		\draw[OIred, ultra thick] plot[domain=-0.99:0.99,samples=400] (\x,{(exp(-1/(1-(\x)*(\x))))});
		\draw[OIred, ultra thick] plot[domain=-1.5:-0.99,samples=400] (\x,0);
		\draw[OIred, ultra thick] plot[domain=0.99:1.5,samples=400] (\x,0);
		\draw[gray] (-0.99,-0.1) -- (-0.99,0.1);
		\draw[gray] (0.99,-0.1) -- (0.99,0.1);
		\node at (-1,-0.3) {\small $-1$} node at (0,-0.3) {\small $0$} node at (1,-0.3) {\small $1$};
		\end{tikzpicture}
	\caption{"Bump" function that is everywhere infinitely often differentiable, but not real analytic.}\label{fig:bump}
\end{wrapfigure}
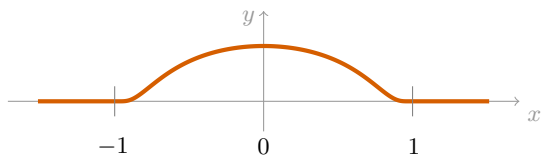

In summary we have found a way to construct features that we can use for the classification problem at hand. But there are some caveats: We have assumed that the functions $f$ are infinitely differentiable and that the remainder vanishes as $n$ goes to zero (we say such a function is \textit{real analytic}). Not all infinitely differentiable functions have this property, e.g., consider the "bump function"
\[
g(x) = \left\{ \begin{array}{ll} e^{-\frac{1}{1-x^2}} & \text{ if } x \in (-1,1) \\ 0 & \text{ else,} \end{array} \right.
\]
which is everywhere infinitely often continuously differentiable, nevertheless the Taylor approximation with a base point outside of the interval $(-1,1)$ will provide as approximation always $0$, at any order. Another point (that we will tackle later) is that our original problem had a discrete time series of data, not continuous functions, so there is no direct way to see how this approach can be put directly to use.

But overall, this approach is promising, and it seems the crucial object we have seen appear are the iterated integrals
\[
I_n(x) = \int_{x_0}^x \int_{x_0}^{x_n} \cdots \int_{x_0}^{x_3} \int_{x_0}^{x_2} 1 \, dx_1 dx_2 \cdots dx_{n-2} dx_{n-1} dx_n.
\]
So far all what these iterated integrals are doing is to calculate the volume $\frac{(x-x_0)^n}{n!}$ of an $n$-dimensional simplex with side length $x - x_0$ as they just integrate $1$ over the $n$-dimensional integration region
\[
\bigl\{ x_0 \leq x_1 \leq x_2 \leq \cdots  \leq x_{n-1} \leq x_n \leq x \bigr\}.
\]
We can, however, build on this concept and mitigate the issues encountered by generalizing the concept of iterated integral. To do so, we have to learn first more about integration.

\section{The Riemann--Stieltjes Integral}

\begin{figure}[htb]
\centering
	\begin{tikzpicture}
		\begin{scope}[scale=0.7]
		\draw[gray!80, ->] (-0.5cm,0cm) -- (8.5cm,0cm) node at (8.3, -0.3) {\small $t$};  
		\draw[gray!80, ->] (0.5cm,-0.5cm) -- (0.5cm,4.5cm) node at (0.3, 4.7) {\small $x$};  
		\node at (1.3cm, -0.4cm) {$a$} node at (6.3cm, -0.4cm) {$b$} node[OIblue, ultra thick] at (7.2cm, 2.4cm) {\textbf{$f$}};
		\draw[OIblue, ultra thick] plot[domain=-0.5:8,samples=400] (\x,{(sin(36*\x)  +3)});
		\foreach \x in {1.3,2.3,...,6.3}
			\draw[black] (\x,0) -- (\x,{(max(sin(36*\x)  +3, sin(36*\x+18)  +3, sin(36*\x-18)  +3)});
		\foreach \x in {1.3,2.3,...,5.8}
			\draw[black] (\x,{(sin(36*\x+18)  +3)}) -- ({(\x + 1)},{(sin(36*\x+18)  +3)});	
		\draw[black, dashed] (4.8,0) -- (4.8,{(sin(36*4.8)  +3)});
		\node at (4.3,-0.3) {\small $t_i$} node at (4.8,-0.3) {\small $\tau_i$} node at (5.4,-0.3) {\small $t_{i+1}$};
		\node[black] at (-0.5,4) {i)};
		\end{scope}
		\begin{scope}[xshift=8.5cm]
		\draw[gray!80, ->] (-1,0,0) node[above] {\small $x$} -- (5,0,0);
 		\draw[gray!80, ->] (0,0,0) -- (0,2.5,0) node[above] {\small $y$};
		\draw[gray!80, ->] (0,0,0) -- (0,0,2) node[left] {\small $z$};
 		\foreach \x in {0.1,0.2,...,3.9}
  		\draw[OIorange] (\x,0,{sin(180*\x)}) -- (\x,{2-0.5*(\x - 2)*(\x-2)},{sin(180*\x)});
		\draw[OIorange, ultra thick] plot[domain=0:4,samples=200] (\x,0,{sin(180*\x)});
 		\draw[OIblue, ultra thick] plot[domain=0:4,samples=200] (\x,{2-0.5*(\x - 2)*(\x-2)},{sin(180*\x)});
		\node[OIorange, ultra thick] at (3.2,0,1) {$g$};
		\node[OIblue, thick] at (3,1.8,0) {$f$};
		\node[black] at (-0.5,2.7,0) {ii)};
\end{scope}
		\end{tikzpicture}
	\caption{i) A Riemann sum approximating the Riemann integral $\int_a^b f(t) \, dt$. ii) The Riemann--Stieltjes integral $\int_a^b f(t) \, dg(t)$ calculates the area under the function $f$ above the path on the $x\,z$-plane described by the function $g$ (here $f(t) = 2\pi-\frac{1}{2\pi}(t-2\pi)^2$ and $g(t) = \sin(t)$ on $[0,4\pi]$).}\label{fig:RS}
\end{figure}
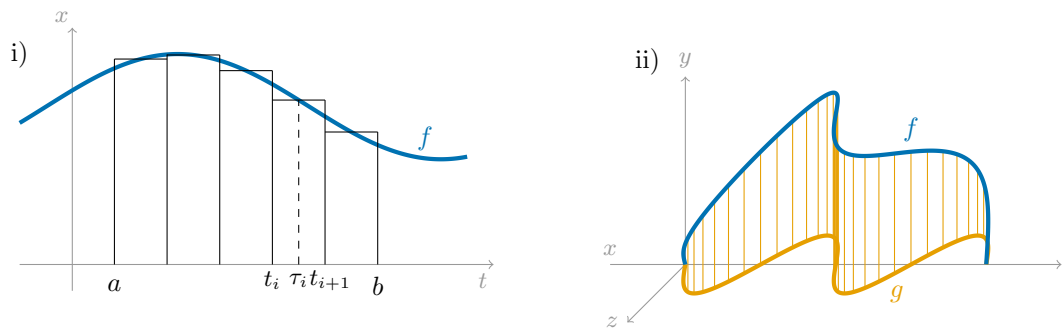

The classical way to introduce the definite integral of a function is as the (signed) area under the function approximated by rectangular areas (cf. Figure \ref{fig:RS} i)), the so-called Riemann sum. As one refines the approximation, the area converges to a limit which is known as the (Riemann) integral. More formally, given a continuous function $f \, : \, [a,b] \to \R$ and a sequence of partitions $\Delta^n$ of the interval $[a,b]$,
\[
\Delta^n := \big\{a = t^n_0 < t^n_1 < \cdots < t^n_{k_n} = b\bigr\}
\]
with mesh converging to zero, i.e.,
\[
\lim_{n \to \infty} \bigl\vert \Delta^n \bigr\vert := \lim_{n \to \infty} \sup_{t^n_i, t^n_{i+1} \in \Delta^n} \bigl\vert t_{i+1}^n - t^n_i \bigr\vert = 0,
\]
we define the Riemann integral as 
\[
\int_a^b f(t) \, dt :=  \lim_{n \to \infty}  \sum_{t^n_i, t^n_{i+1} \in \Delta^n} f(\tau^n_i) \bigl(t^n_{i+1} - t^n_i\bigr)
\]
for arbitrary $\tau^n_i \in \bigl[ t^n_i, t^n_{i+1}\bigr]$. In particular, the value of the integral does not depend on the choice of the $\tau^n_i$s.

But we can apply this approach in the same way much more general: Instead of considering the increments $t^n_{i+1} - t^n_i$ we can first apply a non-decreasing function $g \, :\,  [a,b] \to \R$ to $t$, which yields 
\begin{equation}\label{eq:defRS}
\int_a^b f(t) \, dg(t) :=  \lim_{n \to \infty}  \sum_{t^n_i, t^n_{i+1} \in \Delta^n} f(\tau^n_i) \bigl(g(t^n_{i+1}) - g(t^n_i)\bigr).
\end{equation}
This is known as the \textit{Riemann--Stieltjes integral}. Using the linearity of the sums, we can extend this even more. If $g$ is the difference of two non-decreasing functions $g^+$ and $g^-$, $g(t) = g^+(t) - g^-(t)$ for all $t \in [a,b]$, we can define
\[
\int_a^b f(t) \, dg(t) :=  \int_a^b f(t) \, dg^+(t) - \int_a^b f(t) \, dg^-(t).
\]
It turns out that the functions that can be written as difference of non-decreasing functions are exactly those that are of \textit{bounded (total) variation} (or \textit{BV} functions), satisfying
\[
\TV\nolimits_{a,b}(g) := \sup_{\Delta^n} \sum_{t^n_i, t^n_{i+1} \in \Delta^n} \bigl\vert g(t_{i+1}^n) -g (t^n_i) \bigr\vert < \infty
\]

\begin{wrapfigure}{L}{0.45\textwidth}\centering
	\begin{tikzpicture}[scale=1.2]
		\draw[gray!80, ->] (0cm,-1.5cm) -- (5cm,-1.5cm) node at (4.8, -1.7) {\small $t$};  
		\draw[gray!80, ->] (0.5cm,-2cm) -- (0.5cm,2.5cm) node at (0.3, 2.4) {\small $x$};  
		\draw[OIpink, ultra thick] plot[domain=0.5:4.5,samples=400] (\x,{0.5*(\x-1)*(\x-2)*(\x-4)}) node at (4.2,2) {$g$};
		\draw[black!90, dashed] (0.5,-1.31) -- (1.45,-1.31) (0.5,0.315) -- (3.22, 0.315) (0.5, -1.06) -- (4.5, -1.06);
		\draw[OIgreen, ultra thick] (1.45,-1.31) -- (1.45,0.315) (3.22, 0.315) -- (3.22, -1.06) (4.5, -1.06) -- (4.5, 2.188);
		\end{tikzpicture}
	\caption{The total variation of the function $g$ is given by the total sum of the absolute increments which is the sum of the length of the vertical green lines.}\label{fig:TV}
\end{wrapfigure}
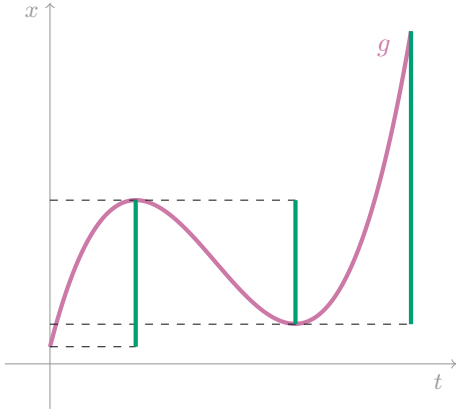

The total variation of a function $g$ measures the total amount of change the function is subjext to. For avid hikers, an easy way to understand this is to think about $g(t)$ as the altitude at a given point $t$ on the route. Then $\TV_{a,b}(g)$ measures the total absolute elevation gain during the hike from $a$ to $b$ (including the height differential made on temporary descents and counter ascents), see Figure \ref{fig:TV}. Thus $BV$ functions are those that correspond to a finite total elevation gain.

From a geometric point of view one can think of $g$ as function in $t$ producing a path (the \textit{graph} of $g$) from the pairs $\bigl(t, g(t)\bigr)$ on the $x$-$z$-plane. The function $f$ is considered a function in the $y$ direction, above the $x$-$z$-plane (depending however only on the $x$-coordinate, $t$ that is). Then the Riemann--Stieltjes integral $\int_a^b f(t) \, dg(t)$ corresponds to the area below the function $f$ \textit{along} the path $g$.

From the definition we can derive two important properties of Riemann--Stieltjes integrals. In the case that $g$ is continuously differentiable we have
\begin{align}\label{eq:idiff}
\int_a^b f(t) \, dg(t) & =  \lim_{n \to \infty}  \sum_{t^n_i, t^n_{i+1} \in \Delta^n} f(\tau^n_i) \bigl(g(t^n_{i+1}) - g(t^n_i)\bigr) \nonumber\\ & =  \lim_{n \to \infty}  \sum_{t^n_i, t^n_{i+1} \in \Delta^n} f(\tau^n_i) \frac{g(t^n_{i+1}) - g(t^n_i)}{t^n_{i+1} - t^n_i}\bigl(t^n_{i+1} - t^n_i\bigr) = \int_a^b f(t)g'(t) \, dt,
\end{align}
so we can express the Riemann-Stieltjes integral as regular Riemann integral, multiplying the integrand with the derivative of $g$. This point of view has an important connection to probability theory and expected values: If $g$ is the cumulative distribution function (aka. cdf) of a continuous random variable $X$ taking values between $a$ and $b$, then $g'$ is the density of this random variable and the formula \eqref{eq:idiff} tells us how to calculate the expectation of $f(X)$.

The second important formula is the generalization of \textit{integration by parts} to Riemann Stieltjes integrals. We observe that by telescopic sums
\begin{align*}
f(b)g(b) - f(a)g(a) &=  \sum_{t^n_i, t^n_{i+1} \in \Delta^n} f(t_{i+1})g(t_{i+1}) - f(t_i)g(t_i) \\
&=  \sum_{t^n_i, t^n_{i+1} \in \Delta^n} f(t_{i+1})g(t_{i+1}) - f(t_{i+1})g(t_i) + f(t_{i+1})g(t_i) - f(t_i)g(t_i)\\
&=  \sum_{t^n_i, t^n_{i+1} \in \Delta^n} f(t_{i+1}) \bigl(g(t_{i+1}) -g(t_i)\bigr) + \sum_{t^n_i, t^n_{i+1} \in \Delta^n} g(t_i) \bigl(f(t_{i+1}) + f(t_i)\bigr),
\end{align*}
and thus, when considering the limit $n \to \infty$, we get
\begin{equation}\label{eq:ibp}
f(b)g(b) - f(a)g(a) = \int_a^b f(t) \, dg(t) + \int_a^b g(t) \, df(t).
\end{equation}
Note that we have here once (namely for $f$) the base points $t_{i+1}$ in the Riemann--Stieltjes sums, and once (for $g$) $t_i$. Note that this does not influence the convergence to the integrals, as we allowed for general basepoints $\tau_i \in [t_i, t_{i+1}]$ in the definition \eqref{eq:defRS}.

Thus, having defined the Riemann--Stieltjes integral, we can think about ways how to improve on the idea of using the Taylor series as tool for feature extraction. An obvious way to do so is to try to use instead of iterated Riemann integrals now iterated Riemann--Stieltjes integrals. We could define
\[
I_n = I_n(x) = \int_{x_0}^x \int_{x_0}^{x_n} \cdots \int_{x_0}^{x_3} \int_{x_0}^{x_2} 1 \, df(x_1) df(x_2) \cdots df(x_{n-2}) df(x_{n-1}) df(x_n).
\]
and hoping the truncated sequence
\[
\bigl(I_0, I_1, I_2, \cdots, I_{n-1}, I_n\bigr).
\]
produces a good feature set. Maybe this is a generalization of the Taylor series, and in the special case of an infinitely often differentiable function we recover the Taylor formulation? Unfortunately, this is not true. We can calculate $I_n$ inductively using integration by parts, which gives us
\begin{equation}\label{eq:simplex}
I_n(x) = \frac{\bigl(f(x)-f(x_0)\bigr)^n}{n!}.
\end{equation}
Indeed, we note first that for all $n$ we have the recursive relation
\[
I_{n+1}(x) = \int_{x_0}^x I_n(y) \, df(y),
\]
from which we can conclude
\begin{align*}
I_{n+1}(x) &= \int_{x_0}^x I_n(x_{n+1}) \, df(x_{n+1}) =  \frac{1}{n!} \int_{x_0}^x \bigl(f(x_{n+1}) - f(x_0)\bigr)^n \, df(x_{n+1})\\
& = \frac{1}{n!} \biggl( \Bigl. \bigl(f(x_{n+1}) - f(x_0)\bigr)^n f(x_{n+1}) \Bigr\vert_{x_{n+1}=x_0}^x - \int_{x_0}^x f(x_{n+1})  \, d\bigl(f(x_{n+1}) - f(x_0)\bigr)^n\biggr)\\
& = \frac{1}{n!} \biggl( \bigl(f(x) - f(x_0)\bigr)^n f(x) - n\int_{x_0}^x f(x_{n+1}) \bigl(f(x_{n+1}) - f(x_0)\bigr)^{n-1}\, df(x_{n+1})\biggr)\\
& = \frac{1}{n!} \biggl( \bigl(f(x) - f(x_0)\bigr)^n f(x) - n \int_{x_0}^x f(x_{n+1}) (n-1)! I_{n-1}(x_{n+1})\, df(x_{n+1})\biggr)\\
& = I_n(x) f(x) - \int_{x_0}^x f(x_{n+1}) I_{n-1}(x_{n+1})\, df(x_{n+1})\\
& = I_n(x) f(x) - \int_{x_0}^x \bigl(f(x_{n+1})-f(x_0)\bigr) I_{n-1}(x_{n+1})\, df(x_{n+1}) - \int_{x_0}^x f(x_{0}) I_{n-1}(x_{n+1})\, df(x_{n+1})\\
& = I_n(x) f(x) - \frac{1}{(n-1)!}\int_{x_0}^x \bigl(f(x_{n+1})-f(x_0)\bigr)^n\, df(x_{n+1}) - f(x_{0}) \int_{x_0}^x  I_{n-1}(x_{n+1})\, df(x_{n+1})\\
\end{align*}
\begin{align*}
& = I_n(x) f(x) - n\int_{x_0}^x I_n(x_{n+1})\, df(x_{n+1}) - f(x_{0}) I_{n}(x)\\
& = I_n(x) f(x) - n I_{n+1}(x) - f(x_0) I_{n}(x)
\end{align*}
and thus
\[
I_{n+1}(x) = \frac{f(x)-f(x_0)}{n+1} I_n(x) = \frac{\bigl(f(x)-f(x_0)\bigr)^{n+1}}{(n+1)!}.
\]
It follows that the truncated sequence depends only on $f$ on the endpoints of the interval $[x_0,x]$, so it does not take into account at all how $f$ behaves inside of the interval. This is disappointing at first -- we will see however that this nevertheless provides the germs of a solution to the problem. But for this we have to first go into higher dimensions and learn about paths.

\section{Paths and their Signatures}

\begin{wrapfigure}{R}{0.6\textwidth}\centering
	\begin{tikzpicture}
		\draw[gray!80, ->] (-0.2cm,0cm) -- (2.3cm,0cm) node at (2.5,0) {\small $x_1$};  
		\draw[gray!80, ->] (0cm,-0.2cm) -- (0cm,4.3cm) node at (0,4.5) {\small $x_2$};  
		\draw[OIgreen, smooth, ultra thick, ->] plot[domain=0:2,samples=400] ({(\x)},{((\x)*(\x))});
		\node at (-0.5cm,4.2cm) {a)} node[OIgreen, thick] at (1.5cm,3.6cm) {$X$};
		\begin{scope}[xshift=6cm, yshift=2cm]
		\draw[gray!80, ->] (-2.3cm,0cm) -- (2.3cm,0cm) node at (2.5,0) {\small $x_1$};  
		\draw[gray!80, ->] (0cm,-2.3cm) -- (0cm,2.3cm) node at (0,2.5) {\small $x_2$};  
		\draw[OIpink, smooth, ultra thick, ->] plot[domain=0:350,samples=400] ({(2*sin(\x))},{(2*cos(\x))});
		\node at (-2.5cm,2.2cm) {b)} node[OIpink, thick] at (1.4cm,2cm) {$Y$};
		\end{scope}
		\end{tikzpicture}
		\caption{Examples of two-dimensional paths.}\label{fig:pathex}
\end{wrapfigure}
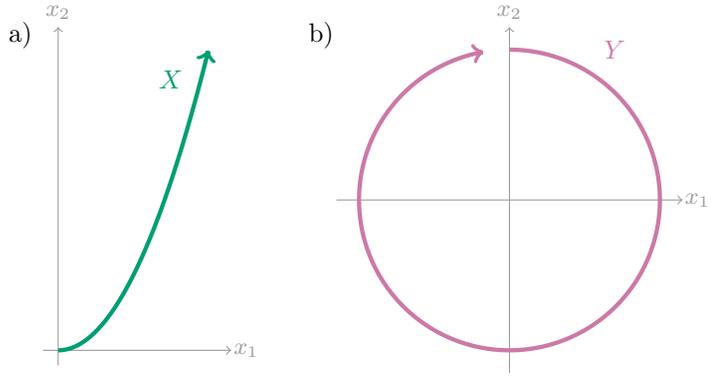

A $d$-dimensional path is a continuous function $X \, : \, [a,b] \to \R^d$, where we will now use the shorthand $X_t$ for $X(t)$. Thus $X_t$ is a $d$-dimensional vector of one-dimensional functions
\[
X_t = \bigl( X^1_t,  X^2_t,  \cdots  X^d_t \bigr).
\]
Simple two-dimensional examples (see Figure \ref{fig:pathex}) would be
\begin{align*}
t & \mapsto (t,t^2)  = \bigl( X^1_t,  X^2_t\bigr) = X_t \\ t & \mapsto \bigl(\sin(t),\cos(t)\bigr)  = \bigl( Y^1_t,  Y^2_t\bigr) = Y_t.
\end{align*}

If a path is of finite total variation, then we can also use its components as integrators for Riemann--Stieltjes integrals. We can use it also for iterated integrals.

Thus, e.g., taking the two-dimensional example $X_t = \bigl( X^1_t,  X^2_t\bigr)$ we can define integrals such as
\[
\int_a^b 1 \, dX^1_{t_1}, \qquad \int_a^b \int_a^{t_2} 1 \, dX^2_{t_1} dX^1_{t_2}, \qquad \int_a^b \int_a^{t_4} \int_a^{t_3}\int_a^{t_2} 1 \, dX^2_{t_1} dX^1_{t_2} dX^2_{t_3} dX^1_{t_4}.
\] 
If we are doing this in a more systematic way, this leads us to the notion of the \textit{signature} of a path.

\begin{definition}
The signature of a $d$-dimensional BV-path $X \, : \, [a,b] \to \R^d$ is the sequence
\begin{equation}\label{eq:defsig} 
\mathcal{S}_{a,b}(X)  = \Bigl( 1, S^1_{a,b}(X), S^2_{a,b}(X), \cdots, S^d_{a,b}(X), S^{1,1}_{a,b}(X), \cdots S^{1,d}_{a,b}(X), S^{2,1}_{a,b}(X), \cdots S^{d,d}_{a,b}(X), S^{1,1,1}_{a,b}(X), \cdots \Bigr)
\end{equation}
where
\[
S^{i_1,i_2,\ldots, i_{n-1}, i_n}_{a,b}(X) = \int_a^b \int_a^{t_n} \cdots \int_a^{t_3}\int_a^{t_2} 1 \, dX^{i_1}_{t_1} dX^{i_2}_{t_2} \cdots dX^{i_{n-1}}_{t_{n-1}} dX^{i_n}_{t_n} \quad \text{ for}\quad  i_1, \ldots, i_n \in \{1,2,\ldots, d\}. 
\]
\end{definition}

To give an example of the construction, let's consider again the path $X_t = \bigl( X^1_t,  X^2_t\bigr) = (t,t^2)$. Then 
\begin{align*}
S^{2,1,2}_{0,1}(X) & = \int_0^1 \int_0^{t_3} \int_0^{t_2}1 \, dX^{2}_{t_1} dX^{1}_{t_2} dX^{2}_{t_3} = \int_0^1 \int_0^{t_3} \int_0^{t_2}1 \, dt_1^2 dt_2 dt_3^2 \\ &= \int_0^1 \int_0^{t_3} \int_0^{t_2}1 \cdot 2t_1 \, dt_1 dt_2 dt_3^2 = \int_0^1 \int_0^{t_3} t_2^2 \, dt_2 dt_3^2 = \int_0^1  \frac{t_3^3}{3} \cdot  2t_3 \, dt_3 = \frac{2}{5}.
\end{align*}

It turns out that the signature provides a very good characterization of a path. Specifically, we have the following theorem, due to Hambly and Lyons \cite{HL10}.

\begin{theorem}\label{thm:uniq}
Let $X \, : \, [a,b] \to \R^d$ be a $d$-dimensional BV-path. Then the signature $\mathcal{S}_{a,b}(X) $ characterizes $X$ uniquely up to \textit{tree-like equivalence}.
\end{theorem} 
The notion of tree-like equivalence is quite a bit involved. But we can gather the essence if we consider reparameterizations. Let $\phi \, : \, [a,b] \to [a,b]$ a non-decreasing continuous function with $\phi(a) =a$ and $\phi(b) = b$. In this case, for each $d$-dimensional path $X_t$, also the path $X_{\phi(t)}$ is a path on the interval $[a,b]$. It turns out that  $X_t$ and $X_{\phi(t)}$ have the same signature as a change of variables $s = \phi(t)$ does not change the (iterated) integrals. This can be understood as that the speed at which a path is run through does not matter for the signature, only its geometry. In the same way one can have that the path is run in some direction for some time backwards and then again forwards -- this part of the path does not contribute to the signature. To provide examples building upon the examples given above: If we reparametrize the path $X \, : \, [0,1] \to \R^2$, $X_t = \bigl(t,t^2\bigr)$ with the function $\phi(t) = t^2$, the iterated integrals do not change. E.g.,
\begin{align*}
\int_0^1 \int_0^{t_2} 1 \, dX^1_{t_1} dX_{t_2}^2 & = \int_0^1 \int_0^{t_2} 1 \, dt_1 dt_2^2 = \int_0^1 t_2 dt_2^2 = \int_0^1 t_2 \cdot 2t_2dt_2 = \frac{2}{3},\\
\int_0^1 \int_0^{t_2} 1 \, dX^1_{\phi(t_1)} dX_{\phi(t_2)}^2 & = \int_0^1 \int_0^{t_2} 1 \, dt_1^2 dt_2^4 = \int_0^1 t_2^2 dt_2^4 = \int_0^1 t^2_2 \cdot 4t_2^3dt_2 = \frac{4}{6} = \frac{2}{3}. 
\end{align*}

\begin{figure}[htb]\centering
	\begin{tikzpicture}
		\draw[gray!80, ->] (-2.3cm,0cm) -- (2.3cm,0cm) node at (2.5,0) {\small $x_1$};  
		\draw[gray!80, ->] (0cm,-2.3cm) -- (0cm,2.3cm) node at (0,2.5) {\small $x_2$};  
		\draw[OIorange, smooth, ultra thick] plot[domain=0:360,samples=400] ({(2*sin(\x))},{(2*cos(\x))});
		\node at (-2.5cm,2.2cm) {i)} node[OIorange, thick] at (1.4cm,2cm) {$Y$};
		\begin{scope}[xshift=5.5cm]
		\draw[gray!80, ->] (-2.3cm,0cm) -- (2.3cm,0cm) node at (2.5,0) {\small $x_1$};  
		\draw[gray!80, ->] (0cm,-2.3cm) -- (0cm,2.3cm) node at (0,2.5) {\small $x_2$};  
		\draw[OIbluelight, smooth, ultra thick] plot[domain=0:360,samples=400] ({(2*sin(\x))},{(2*cos(\x))});
		\draw[OIbluelight, smooth, ultra thick]  (1.41cm,1.41cm) -- (1.81cm,1.81cm);
		\node at (-2.5cm,2.2cm) {ii)} node[OIbluelight, thick] at (1.4cm,2cm) {$\tilde{Y}$};
		\end{scope}
		\begin{scope}[xshift=11cm]
		\node at (-2.5cm,2.2cm) {iii)};
		\draw[gray!80, ->] (-2.3cm,0cm) -- (2.3cm,0cm) node at (2.5,0) {\small $x_1$};  
		\draw[gray!80, ->] (0cm,-2.3cm) -- (0cm,2.3cm) node at (0,2.5) {\small $x_2$};  
		\draw[OIgreen, smooth, ultra thick] plot[domain=0:360,samples=400] ({(2*sin(\x))},{(2*cos(\x))});
		\draw[OIgreen, smooth, ultra thick]  (1.41cm,1.41cm) -- (2.21cm,2.21cm);
		\draw[OIgreen, smooth, ultra thick]  (2.15cm,2.15cm) -- (2cm,2.15cm);
		\draw[OIgreen, smooth, ultra thick]  (2.15cm,2.15cm) -- (2.15cm,2cm);
		\draw[OIgreen, smooth, ultra thick]  (2cm,2cm) -- (1.7cm,2cm);
		\draw[OIgreen, smooth, ultra thick]  (2cm,2cm) -- (2cm,1.7cm);
		\draw[OIgreen, smooth, ultra thick]  (1.8cm,1.8cm) -- (1.4cm,1.8cm);
		\draw[OIgreen, smooth, ultra thick]  (1.8cm,1.8cm) -- (1.8cm,1.4cm);
		\begin{scope}[rotate=18]
		\draw[OIgreen, smooth, ultra thick]  (1.41cm,1.41cm) -- (2.21cm,2.21cm);
		\draw[OIgreen, smooth, ultra thick]  (2.15cm,2.15cm) -- (2cm,2.15cm);
		\draw[OIgreen, smooth, ultra thick]  (2.15cm,2.15cm) -- (2.15cm,2cm);
		\draw[OIgreen, smooth, ultra thick]  (2cm,2cm) -- (1.7cm,2cm);
		\draw[OIgreen, smooth, ultra thick]  (2cm,2cm) -- (2cm,1.7cm);
		\draw[OIgreen, smooth, ultra thick]  (1.8cm,1.8cm) -- (1.4cm,1.8cm);
		\draw[OIgreen, smooth, ultra thick]  (1.8cm,1.8cm) -- (1.8cm,1.4cm);
		\end{scope}
		\begin{scope}[rotate=36]
		\draw[OIgreen, smooth, ultra thick]  (1.41cm,1.41cm) -- (2.21cm,2.21cm);
		\draw[OIgreen, smooth, ultra thick]  (2.15cm,2.15cm) -- (2cm,2.15cm);
		\draw[OIgreen, smooth, ultra thick]  (2.15cm,2.15cm) -- (2.15cm,2cm);
		\draw[OIgreen, smooth, ultra thick]  (2cm,2cm) -- (1.7cm,2cm);
		\draw[OIgreen, smooth, ultra thick]  (2cm,2cm) -- (2cm,1.7cm);
		\draw[OIgreen, smooth, ultra thick]  (1.8cm,1.8cm) -- (1.4cm,1.8cm);
		\draw[OIgreen, smooth, ultra thick]  (1.8cm,1.8cm) -- (1.8cm,1.4cm);
		\end{scope}
		\begin{scope}[rotate=54]
		\draw[OIgreen, smooth, ultra thick]  (1.41cm,1.41cm) -- (2.21cm,2.21cm);
		\draw[OIgreen, smooth, ultra thick]  (2.15cm,2.15cm) -- (2cm,2.15cm);
		\draw[OIgreen, smooth, ultra thick]  (2.15cm,2.15cm) -- (2.15cm,2cm);
		\draw[OIgreen, smooth, ultra thick]  (2cm,2cm) -- (1.7cm,2cm);
		\draw[OIgreen, smooth, ultra thick]  (2cm,2cm) -- (2cm,1.7cm);
		\draw[OIgreen, smooth, ultra thick]  (1.8cm,1.8cm) -- (1.4cm,1.8cm);
		\draw[OIgreen, smooth, ultra thick]  (1.8cm,1.8cm) -- (1.8cm,1.4cm);
		\end{scope}
		\begin{scope}[rotate=72]
		\draw[OIgreen, smooth, ultra thick]  (1.41cm,1.41cm) -- (2.21cm,2.21cm);
		\draw[OIgreen, smooth, ultra thick]  (2.15cm,2.15cm) -- (2cm,2.15cm);
		\draw[OIgreen, smooth, ultra thick]  (2.15cm,2.15cm) -- (2.15cm,2cm);
		\draw[OIgreen, smooth, ultra thick]  (2cm,2cm) -- (1.7cm,2cm);
		\draw[OIgreen, smooth, ultra thick]  (2cm,2cm) -- (2cm,1.7cm);
		\draw[OIgreen, smooth, ultra thick]  (1.8cm,1.8cm) -- (1.4cm,1.8cm);
		\draw[OIgreen, smooth, ultra thick]  (1.8cm,1.8cm) -- (1.8cm,1.4cm);
		\end{scope}
		\begin{scope}[rotate=90]
		\draw[OIgreen, smooth, ultra thick]  (1.41cm,1.41cm) -- (2.21cm,2.21cm);
		\draw[OIgreen, smooth, ultra thick]  (2.15cm,2.15cm) -- (2cm,2.15cm);
		\draw[OIgreen, smooth, ultra thick]  (2.15cm,2.15cm) -- (2.15cm,2cm);
		\draw[OIgreen, smooth, ultra thick]  (2cm,2cm) -- (1.7cm,2cm);
		\draw[OIgreen, smooth, ultra thick]  (2cm,2cm) -- (2cm,1.7cm);
		\draw[OIgreen, smooth, ultra thick]  (1.8cm,1.8cm) -- (1.4cm,1.8cm);
		\draw[OIgreen, smooth, ultra thick]  (1.8cm,1.8cm) -- (1.8cm,1.4cm);
		\end{scope}
		\end{scope}
	\end{tikzpicture}
	\caption{i) Path $Y$. ii) Path $\tilde{Y}$ that has same signature as $Y$ as it is tree-like equivalent. iii) Path even more tree-like (forest-like?) equivalent to $Y$, so signatures are indistinguishable}\label{fig:tree}
\end{figure}
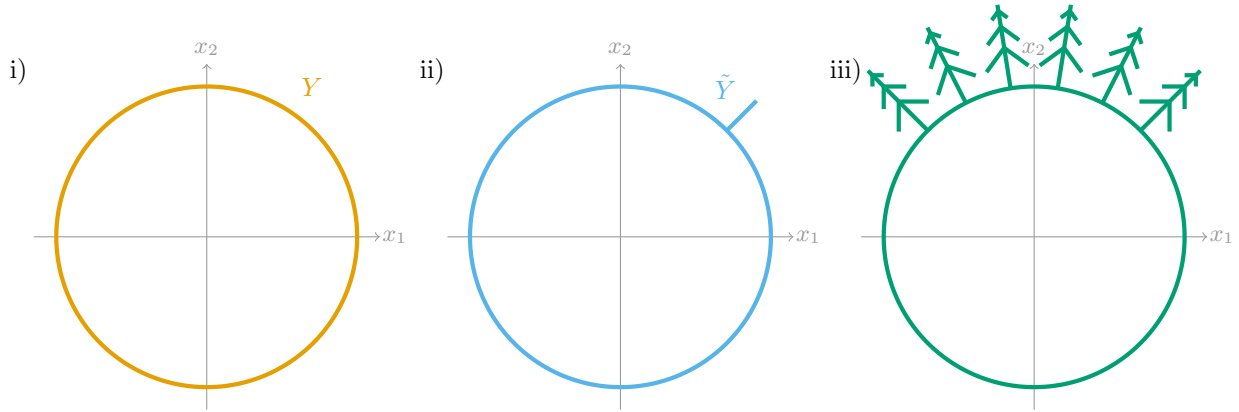

Considering the path $Y \, : \, [0,1] \to \R^2$, $Y_t  = \bigl(\sin(t),\cos(t)\bigr)$, we note that the path $\tilde{Y} \, : \, [0,1] \to \R^2$ given by
\[
\tilde{Y}_t = \left \{ \begin{array}{ll} \bigl(\sin(2t), \cos(2t)\bigr) & \text { if } t \in \bigl[0, \frac{\pi}{8}\bigr], \\ \Bigl(\frac{\sqrt{2}}{2} + \bigl(t-\frac{\pi}{8}\bigr), \frac{\sqrt{2}}{2} + \bigl(t-\frac{\pi}{8}\bigr)\Bigr) & \text { if } t \in \bigl(\frac{\pi}{8}, \frac{3\pi}{16}\bigr], \\
\Bigl(\frac{\sqrt{2}}{2} + \bigl(\frac{3\pi}{16} -t\bigr), \frac{\sqrt{2}}{2} + \bigl(\frac{3\pi}{16} - t\bigr)\Bigr) & \text { if } t \in \bigl(\frac{3\pi}{16}, \frac{\pi}{4} \bigr], \\ \bigl(\sin(t), \cos(t)\bigr) & \text { if } t \in \bigl(\frac{\pi}{4}, 2\pi\bigr] \end{array}\right.
\]
has the same signature as $Y$, see Figure \ref{fig:tree}.

We want also reconnect to the end of the previous section. There we have seen that the iterated integrals alone (which we can now understand as signature of a one-dimensional path) were not really useful to capture the properties of a function $f$. This seems to be in contradiction to Theorem \ref{thm:uniq}. But on closer inspection it isn't, it is rather the issue that one-dimensional setting is a special case in which the result degenerates. Every non-decreasing function $f \, : \, [a,b] \to \R$, $f(a) = a$, $f(b) = b$ is just a reparameterization if its image completely contained in the interval $[a,b]$. If $f$ is not non-decreasing, it has some parts through it runs both, forwards and backwards, thus producing no additional contribution to the signature. So, the signature is indeed completely determined by $f(a)$ and $f(b)$ - all functions with these start- and end values are tree-like equivalent. But this does not mean that the signature is not useful to recover the function $f$, one has just to go into higher dimensions. E.g., one can consider the \textit{graph} of $f$ by considering the two-dimensional path $X \, : \, [a,b] \to \R^2$, $X_t = \bigl(t, f(t)\bigr)$. The signature $\mathcal{S}_{a,b}(X)$ now indeed characterizes the function $f$ uniquely, as the first dimension acts as auxiliary variable and essentially prescribes the parameterization and makes sure that no forth-back excursion is happening. This points to a general principle: every signature can be made a unique description of a path as long as we add to the path an additional dimension, just capturing time, before calculating the signature.
 
Practically we cannot use the full signature for classification, we will truncate it, using only iterated integrals up to order $N$, 
\begin{equation}\label{eq:deftrincsig} 
\mathcal{S}^N_{a,b}(X) = \Bigl( 1, S^1_{a,b}(X), S^2_{a,b}(X), \cdots, S^d_{a,b}(X),  S^{1,1}_{a,b}(X),  \cdots  S^{d,d}_{a,b}(X), S^{1,1,1}_{a,b}(X), \cdots, S^{\overset{N\, \text{times}}{\overbrace{d,d, \cdots,d}}}_{a,b}(X) \Bigr)
\end{equation}

To consider a specific example, we can calculate the $3$-rd level signature of the path $X \, : \, [0,1] \to \R^2$, $X_t = (X^1_t,X^2_t) = (t^2, t^3-t)$:
\begin{align*}
\mathcal{S}^3_{0,1}(X) & = \Biggl( 1, \int_0^1 1 \, dX^1_{t_1}, \int_0^1 1 \, dX^2_{t_1}, \int_0^1 \int_0^{t_2} 1 \, dX^1_{t_1} dX^1_{t_2},  \int_0^1 \int_0^{t_2} 1 \, dX^1_{t_1} dX^2_{t_2}, \int_0^1 \int_0^{t_2} 1 \, dX^2_{t_1} dX^1_{t_2}, \Bigr. \\ & \phantom{==} \int_0^1 \int_0^{t_2} 1 \, dX^2_{t_1} dX^2_{t_2}, \int_0^1 \int_0^{t_3} \int_0^{t_2} 1 \, dX^1_{t_1}  dX^1_{t_2} dX^1_{t_3},  \int_0^1 \int_0^{t_3} \int_0^{t_2} 1 \, dX^1_{t_1}  dX^1_{t_2} dX^2_{t_3}, \\ & \phantom{==}  \int_0^1 \int_0^{t_3} \int_0^{t_2} 1 \, dX^1_{t_1}  dX^2_{t_2} dX^1_{t_3},  \int_0^1 \int_0^{t_3} \int_0^{t_2} 1 \, dX^1_{t_1}  dX^2_{t_2} dX^2_{t_3},  \int_0^1 \int_0^{t_3} \int_0^{t_2} 1 \, dX^2_{t_1}  dX^1_{t_2} dX^1_{t_3}, \\ &\phantom{==}  \Bigl. \int_0^1 \int_0^{t_3} \int_0^{t_2} 1 \, dX^2_{t_1}  dX^1_{t_2} dX^2_{t_3},  \int_0^1 \int_0^{t_3} \int_0^{t_2} 1 \, dX^2_{t_1}  dX^2_{t_2} dX^1_{t_3},  \int_0^1 \int_0^{t_3} \int_0^{t_2} 1 \, dX^2_{t_1}  dX^2_{t_2} dX^2_{t_3}\Biggr)\\
& = \Bigl( 1, 1, 0, \frac{1}{2}, \frac{4}{15}, -\frac{4}{15}, 0, \frac{1}{6}, \frac{4}{35}, \frac{4}{105}, \frac{1}{24}, -\frac{16}{105}, -\frac{1}{12}, \frac{1}{24}, 0 \Bigr). 
\end{align*}
Here we have conducted the calculation stepwise and used the formula for Riemann-Stieltjes integrals with differentiable integrator, e.g.,
\begin{align*}
\int_0^1 \int_0^{t_3} \int_0^{t_2} 1 \, dX^2_{t_1}  dX^1_{t_2} dX^2_{t_3} & = \int_0^1 \int_0^{t_3} \int_0^{t_2} 1 \, d(t_1^3-t_1)  dt_2^2 d(t_3^3-t_3) = \int_0^1 \int_0^{t_3} t_2^3-t_2 \,  dt_2^2 d(t_3^3-t_3) \\
& = \int_0^1 \int_0^{t_3} \bigl(t_2^3-t_2\bigr) 2 t_2 \, dt_2 d(t_3^3-t_3) = \int_0^1 \frac{2}{5} t_3^5-\frac{2}{3}t_3^3 \, d(t_3^3-t_3) \\ &= \int_0^1 \Bigl(\frac{2}{5} t_3^5-\frac{2}{3}t_3^3 \Bigr) \cdot \bigl( 3t_3^2 -1\bigr)\, dt_1 = \frac{3}{20} - \frac{2}{5} + \frac{1}{6} = -\frac{1}{12}.
\end{align*}

\begin{figure}[!htb]
\centering
	\begin{tikzpicture}
		\begin{scope}[scale=0.8]
		\node at (-1cm,6cm) {i)};
		\draw[gray!80, ->] (-0.5cm,0cm) -- (8.5cm,0cm) node at (8.7, -0.3) {\small $x_1$};  
		\draw[gray!80, ->] (0cm,-0.5cm) -- (0cm,6cm) node at (-0.4, 6) {\small $x_2$};  
		\draw (1,1) -- (1,5) -- (7,5);
		\draw (1,1) -- (7,1) -- (7,5);
		\fill[OIblue!20] (1,1) -- plot[smooth, tension=0.8] coordinates {(1,1) (1.5,1.3) (2,1.5) (2.5,3) (3,4) (3.5,3.5) (4,3) (4.5,3.5) (5,4) (5.5,4.5) (6,3.5) (6.5, 4.5) (7,5)} -- (7,1) -- cycle;
		\fill[OIorange!20] (1,1) -- plot[smooth, tension=0.8] coordinates {(1,1) (1.5,1.3) (2,1.5) (2.5,3) (3,4) (3.5,3.5) (4,3) (4.5,3.5) (5,4) (5.5,4.5) (6,3.5) (6.5, 4.5) (7,5)} -- (1,5) -- cycle;
		\draw[OIblue, ultra thick] plot[smooth, tension=0.8] coordinates {(1,1) (1.5,1.3) (2,1.5) (2.5,3) (3,4) (3.5,3.5) (4,3) (4.5,3.5) (5,4) (5.5,4.5) (6,3.5) (6.5, 4.5) (7,5)};
		\node[OIblue] at (4.5,2) {\small $\int_a^b \int_a^{t_2} 1 \, dX^2_{t_1} dX^1_{t_2}$} node[OIred] at (3,4.5) {\small $\int_a^b \int_a^{t_2} 1 \, dX^1_{t_1} dX^2_{t_2}$};
		\draw[black!90, dashed] (0,5) -- (1,5) (0,1) -- (1,1) (1,0) -- (1,1) (7,0) -- (7,1) node at (-0.5,1) {\small $X^2_a$} node at (-0.5,5) {\small $X^2_b$} node at (1,-0.5) {\small $X^1_a$} node at (7,-0.5) {\small $X^1_b$};
		\end{scope}
		\begin{scope}[xshift=8cm, scale=0.8]
		\node at (-1cm,6cm) {ii)};
		\draw[gray!80, ->] (-0.5cm,0cm) -- (8.5cm,0cm) node at (8.7, -0.3) {\small $x_1$};  
		\draw[gray!80, ->] (0cm,-0.5cm) -- (0cm,6cm) node at (-0.4, 6) {$x_2$};  
		\fill[OIblue!20] (1,1) -- plot[smooth, tension=0.8] coordinates {(1,1) (1.5,2) (2,2.5) (2.5,3)} -- (7,3) -- (7,1) -- cycle;
		\fill[OIorange!20] (1,1) -- plot[smooth, tension=0.8] coordinates {(1,1) (1.5,2) (2,2.5) (2.5,3)} -- (1,3) -- cycle;
		\fill[OIpink!20] (2.5,3) -- plot[smooth, tension=0.8] coordinates {(2.5,3) (3,3.5) (3.5,4) (4,5) (4.5,5.5) (5,5.5) (5.5,5) (6,4.5) (6.5, 5) (7,5.5) (7.5,5) (8,4.5) (8.5, 4) (7.5,3.5) (7,3)} -- cycle;
		\draw[OIblue, ultra thick] plot[smooth, tension=0.8] coordinates {(1,1) (1.5,2) (2,2.5) (2.5,3) (3,3.5) (3.5,4) (4,5) (4.5,5.5) (5,5.5) (5.5,5) (6,4.5) (6.5, 5) (7,5.5) (7.5,5) (8,4.5) (8.5, 4) (7.5,3.5) (7,3)};
		\draw (1,1) -- (1,3) -- (7,3);
		\draw (1,1) -- (7,1) -- (7,3);
		\draw[black!90, dashed] (0,3) -- (1,3) (0,1) -- (1,1) (1,0) -- (1,1) (7,0) -- (7,1) node at (-0.5,1) {\small $X^2_a$} node at (-0.5,3) {\small $X^2_b$} node at (1,-0.3) {\small $X^1_a$} node at (7,-0.5) {\small $X^1_b$};
		\node[OIblue] at (4.5,2) {\Large $+$} node[OIred] at (1.5,2.5) {\Large $+$} node at (5.5,3.8) {\Large {\color{OIblue}$+$}  / {\color{OIred}$-$}};
		\end{scope}
	\end{tikzpicture}
	\caption{Geometric interpretation of the mixed second order terms of the signature. i) If the path stays in the rectangle it spans, it decomposes the rectangle in the two signature parts. ii) If the path leaves the square, additional areas have to be counted with signs. Here $\int_a^b \int_a^{t_2} 1 \, dX^2_{t_1} dX^1_{t_2}$ is the sum of the blue and pink areas while $\int_a^b \int_a^{t_2} 1 \, dX^1_{t_1} dX^2_{t_2}$ is the difference between the orange and the pink.}\label{fig:area}
\end{figure}
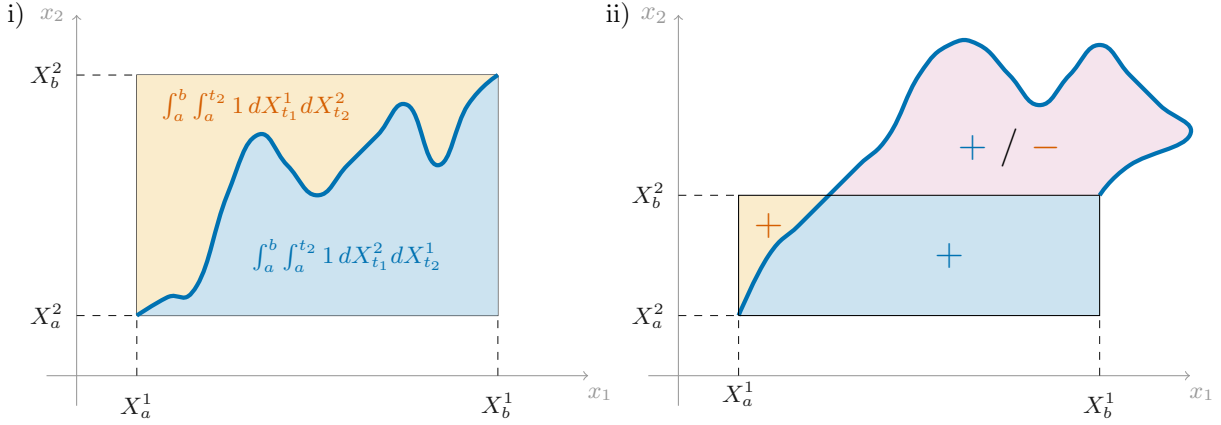

While there is in general not an easy way to illustrate signatures, there is a useful way to picture the two-dimensional case. Starting with the case where both integrals are taken with respect to the same path component, we note that the formula \eqref{eq:simplex} identifies them as (signed) volume of the $n$-dimensional simplex with side length $\vert X^i_b - X^i_a\vert$. We complement this by looking at the mixed integral terms: if we picture (see Figure \ref{fig:area}) the path in the two spatial coordinates, the two mixed integrals $\int_a^b \int_a^{t_1} 1\,  dX_{t_1}^1 dX_{t_1}^2$ and $\int_a^b \int_a^{t_1} 1\,  dX_{t_1}^2 dX_{t_1}^1$ can be seen as the area under the curve resp. the area left of the curve (i.e., below the curve when axes are flipped). This is straight forward as long as the path stays in the rectangle spanned and has no self-intersections. Otherwise, one has to carefully consider which parts are circled clockwise or counterclockwise and make their sign dependent on it - this generalizes the interpretation of the one-dimensional integrals where areas below the $x$-axis are counted negatively. We note also that this give a geometric interpretation of the integration by parts formula \eqref{eq:ibp}:
As 
\[
\int_a^b \int_a^{t_2} 1\,  dX_{t_1}^1 dX_{t_2}^2 + \int_a^b \int_a^{t_2} 1\,  dX_{t_1}^2 dX_{t_2}^1 = \bigl( X^1_b - X^1_a\bigr)\bigl(X^2_b - X^2_a\bigr)
\]
we have
\begin{align*}
\int_a^b X_{t_1}^1 \, dX_{t_1}^2 + \int_a^b X_{t_1}^2 \,  dX_{t_1}^1 & = \int_a^b \int_a^{t_2} 1\,  dX_{t_1}^1 dX_{t_2}^2 + X^1_a(X^2_b - X^2_a) + \int_a^b \int_a^{t_2} 1\,  dX_{t_1}^2 dX_{t_2}^1 + X^2_a(X^1_b - X^1_a) \\ 
&= \bigl( X^1_b - X^1_a\bigr)\bigl(X^2_b - X^2_a\bigr) + X^1_a(X^2_b - X^2_a)+ X^1_a(X^2_b - X^2_a) = X^1_bX^2_b - X^1_aX^2_a.
\end{align*}

\begin{wrapfigure}{R}{0.45\textwidth}\centering
	\begin{tikzpicture}[scale=1.2]
		\draw[gray!80, ->] (-0.5cm,0cm) -- (5cm,0cm)  node at (5.2, -0.2) {\small $x_1$};  
		\draw[gray!80, ->] (0cm,-0.5cm) -- (0cm,2cm)  node at (-0.2, 2.2) {\small $x_2$};  
		\draw[OIred, ultra thick, ->] plot[domain=0:4.5,samples=400] (\x,{0.05*(\x)*(\x+3)}) node at (4,2) {$X_t = (X^1_t,X^2_t)$};
		\draw[black!90, dashed] (2,1.4) -- (2,0.5) (4,1.4) -- (2, 1.4) node(not) at (2,1.4) {$\bullet$} node[above left =-0.2cm of not] {$(x_1,x_2)$} ;
		\node[OIred] at (2,0.5) {$\bullet$} node[OIred] at (4,1.4) {$\bullet$} node[OIred] at (2.3,0.3) {$X_{t_1}$} node[OIred] at (4.3,1.1) {$X_{t_2}$};
		\end{tikzpicture}
	\caption{The geometric interpretation of the iterated integrals. The point $(x_1,x_2)$ is above the path as the path passes first at $t_1$ the length of $x_1$ and only later at $t_2$ the height of $x_2$.}\label{fig:points}
\end{wrapfigure}
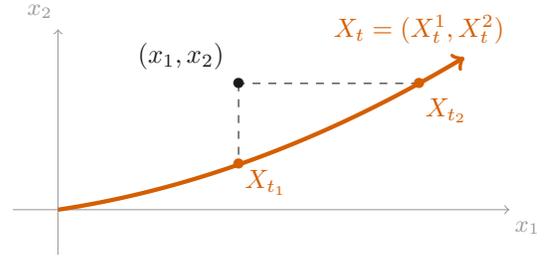

To understand this better and enable us to generalize the understanding to higher dimensions, let's focus first on paths that are increasing in every component (we relegate the discussion of the general case to Appendix \ref{sec:Chen}. Then, in two dimensions, we can decide if a point $(x_1,x_2)$ lies in the area above or below the path by considering if the path hits first the horizontal level of $x_1$ or the vertical level of $x_2$. In the former case the point is above the path, in the latter below. More formally we can introduce the component-wise inverse of the path $X$, $X_i^{-1}(x_i) : \, [X^i(a), X^i(b)] \to [a,b]$. In the two dimensional case we have then the points above the path characterized as $X_1^{-1}(x_1) < X_2^{-1}(x_2)$ whereas for the points below the path satisfy the reverse inequality. Specifically, we see that the areas spanned by the iterated integrals are
\begin{align*}
\int_a^b \int_a^{t_1} 1\,  dX_{t_1}^1 dX_{t_1}^2 &= \ar \Bigl(\bigl\{ (x_1, x_2) \in [X^1(a), X^1(b)] \times [X^2(a), X^2(b)] \, : \,  X_1^{-1}(x_1) < X_2^{-1}(x_2)\bigr\}\Bigr),\\
\int_a^b \int_a^{t_1} 1\,  dX_{t_1}^2 dX_{t_1}^1 &= \ar \Bigl(\bigl\{ (x_1, x_2) \in [X^1(a), X^1(b)] \times [X^2(a), X^2(b)] \, : \,  X_1^{-1}(x_1) > X_2^{-1}(x_2)\bigr\}\Bigr).
\end{align*}
But this can now be generalized into higher dimensions. For the three-dimensional case we have
\begin{align*}
& \phantom{=:}\int_a^b \int_a^{t_3} \int_a^{t_2} 1\,  dX_{t_1}^{i_1} dX_{t_2}^2{i_2} dX_{t_3}^{i_3} \\
& = \vol \biggl(\Bigl\{ (x_1, x_2, x_3 ) \in \prod_{k=1}^3 [X^{i_k}(a), X^{i_k}(b)]  \, : \,  X_{i_1}^{-1}(x_1) < X_{i_2}^{-1}(x_2) < X_{i_3}^{-1}(x_3)\Bigr\}\biggr)
\end{align*}
In the case that all three components are different, this means that the 6 different iterated integrals we get by permuting the integrators correspond to different volumina which together make up the cuboid spanned by the path. Conversely, the cuboid $\prod_{k=1}^3 [X^{i_k}(a), X^{i_k}(b)]$ is cut by the three deformed planes $\{(x_1, x_2, x_3 ) \, : X_{i_1}^{-1}(x_1) = X_{i_2}^{-1}(x_2)\}$, $\{(x_1, x_2, x_3 ) \, : X_{i_1}^{-1}(x_1) = X_{i_3}^{-1}(x_3)\}$ and $\{(x_1, x_2, x_3 ) \, : X_{i_2}^{-1}(x_2) = X_{i_3}^{-1}(x_3)\}$ into $6$ pieces, each of them has a volume calculated by one of the $6$ iterated integrals. In the general case we will have $n!$ (hyper)volumes that add up to the $n$-dimensional hyperrectangle. In the case that all integrators are the same component, the iterated integral corresponds to an $n$-dimensional simplex spanned by the edges of the hyperrectangle.

\section{Data Streams}\label{sec:data}

So far our discussion was about characterizing functions, but we don't want to lose our original goal, to classify time series data, out of sight. The data we are given are a \textit{data stream}, i.e., a collection of data points that form a multi-dimensional time series. In recent literature the terminology \textit{data stream} is preferred to highlight the online and continuing nature of data, formally there is no difference to a classical (multi-dimensional time series). Formally a data stream is a family of tuples
\[
\Bigl( t, D(t) \Bigr)_{ t \in (t_0, t_1, \ldots, t_n)}
\]
where $t$ is (one-dimensional) time and $D(t) \in \mathbb{R}^d$ is the $d$-dimensional data point at time $t$.

To make the data stream amenable to our theory of signatures, we have first to construct continuous paths out of the data stream. This can be done in many different ways. Maybe the most straightforward idea, namely holding the path constant between the times at which we have data, does not work as the path constructed in this way is not continuous. But one easy way to construct a path is to interpolate linearly,
\[
X_t = D(t_i) + (t -t_i) D(t_{i+1}) \quad \text{ for } t \in \bigl[t_i, t_{i+1}\bigr).
\]
We can also add the time as explicit component and construct the two-dimensional path $\tilde{X}_t = (t, X_t)$. This is commonly known as the \textit{time-augmented} path. But there are many more ways, e.g., one can get smoother paths by using piecewise quadratic or higher order polynomials. One could also consider a function that is constant for some time before interpolating to the next data point. Even the piecewise constant interpolation can get made work, albeit only in the time-augmented version:
\[
Y_t = \left\{ \begin{array}{ll} \bigl(t_i + 2(t -t_i), D(t_i) \bigr)  & \quad \text{ for } t \in \bigl[t_i, \frac{t_{i+1}+t_i}{2}\bigr), \\
\bigl(t_{i+1}, D(t_i) + \frac{2t - t_{i+1}-t_i}{t_{i+1}-t_i}\bigl(D(t_{i+1})-D(t_i)\bigr)  & \quad \text{ for } t \in \bigl[\frac{t_{i+1}+t_i}{2}, t_{i+1}\bigr). \end{array} \right.
\]

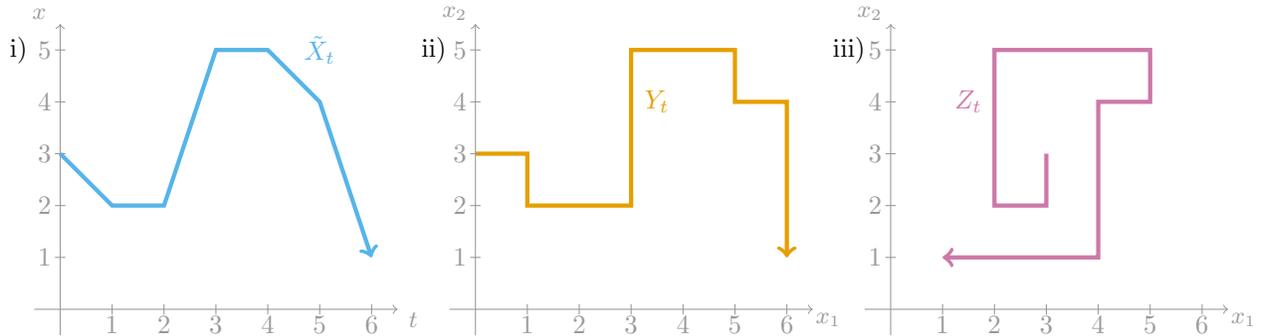
\begin{figure}[htb]
\centering
	\begin{tikzpicture}[scale=0.69]
		\draw[gray!80, ->] (-0.5cm,0cm) -- (6.5cm,0cm) node at (6.8, -0.2) {\small $t$};  
		\draw[gray!80, ->] (0cm,-0.5cm) -- (0cm,5.5cm) node at (-0.4, 5.7) {\small $x$};  
		\draw[OIbluelight, ultra thick, ->] (0,3) -- (1,2) -- (2,2) -- (3,5) -- (4,5) -- (5,4) -- (6,1);
		\foreach \x in {1,2,...,6}
			\draw[gray!80] (\x,-0.1) -- (\x,0.1) node at (\x,-0.3) {$\x$};
		\foreach \x in {1,2,...,5}
			\draw[gray!80] (-0.1,\x) -- (0.1,\x) node at (-0.3,\x) {$\x$};
		\node at (-0.8,5) {i)} node[OIbluelight, thick] at (5,5) {$\tilde{X}_t$};
		\begin{scope}[xshift=8cm]
		\draw[gray!80, ->] (-0.5cm,0cm) -- (6.5cm,0cm) node at (6.8, -0.2) {\small $x_1$};  
		\draw[gray!80, ->] (0cm,-0.5cm) -- (0cm,5.5cm) node at (-0.4, 5.7) {\small $x_2$};  
		\draw[OIorange, ultra thick, ->] (0,3) -- (1,3) -- (1,2) -- (2,2) -- (3,2) -- (3,5) -- (4,5) -- (4,5) -- (5,5) -- (5,4) -- (6,4) -- (6,1);
		\foreach \x in {1,2,...,6}
			\draw[gray!80] (\x,-0.1) -- (\x,0.1) node at (\x,-0.3) {$\x$};
		\foreach \x in {1,2,...,5}
			\draw[gray!80] (-0.1,\x) -- (0.1,\x) node at (-0.3,\x) {$\x$};
		\node at (-0.8,5) {ii)} node[OIorange, thick] at (3.5,4) {$Y_t$};
		\end{scope}
		\begin{scope}[xshift=16cm]
		\draw[gray!80, ->] (-0.5cm,0cm) -- (6.5cm,0cm) node at (6.8, -0.2) {\small $x_1$};  
		\draw[gray!80, ->] (0cm,-0.5cm) -- (0cm,5.5cm) node at (-0.4, 5.7) {\small $x_2$};  
		\draw[OIpink, ultra thick, ->] (3,3) -- (3,2) -- (2,2) -- (2,2) -- (2,2) -- (2,5) -- (5,5) -- (5,5) -- (5,5) -- (5,4) -- (4,4) -- (4,1) -- (1,1);
		\foreach \x in {1,2,...,6}
			\draw[gray!80] (\x,-0.1) -- (\x,0.1) node at (\x,-0.3) {$\x$};
		\foreach \x in {1,2,...,5}
			\draw[gray!80] (-0.1,\x) -- (0.1,\x) node at (-0.3,\x) {$\x$};
		\node at (-0.8,5) {iii)} node[OIpink, thick] at (1.5,4) {$Z_t$};
		\end{scope}
	\end{tikzpicture}
	\caption{Different path constructions from the time series $(3,2,2,5,5,4,1)$. i) Piecewise linear time-augmented, ii) piecewise constant, iii) piecewise constant lead-lag.}\label{fig:constr}
\end{figure}

But there are other ideas to capture additional information by increasing the dimensionality. One is intended for the case where different components of the time series represent different indicators, some leading and some lagging. To get a handle on this dynamics one can consider the path given by the \textit{lead-lag transform}. One doubles the dimensionality of the data stream and shifts one dimension in time (which also nearly doubles the length of the data stream; we can rescale if needed). One possible implementation is
\[
\tilde{D}(t_k) =  \left\{ \begin{array}{ll} \bigl(D(t_i), D(t_{i})\bigr)  & \quad \text{ if } k = 2i, \\
\bigl(D(t_i), D(t_{i+1})\bigr)  & \quad \text{ if } k = 2i +1, \end{array} \right.
\]
for all $t_k \in \{t_0, \ldots, t_{2n}\}$. Then one can use any of the aforementioned methods to produce a path, now from the enlarged data stream $\bigl(t, \tilde{D}(t)\bigr)$. E.g., applying to $(3,2,2,5,5,4,1)$ the lead-lag transform gives the data stream
\[
\bigl( (3,3), (3,2), (2,2), (2,2), (2,2), (2,5), (5,5), (5,5), (5,5), (5,4), (4,4), (4,1), (1,1) \bigr)
\]
which leads in a piecewise constant interpolation to the path depicted in Figure \ref{fig:constr} iii).

\section{Conclusion and Outlook}

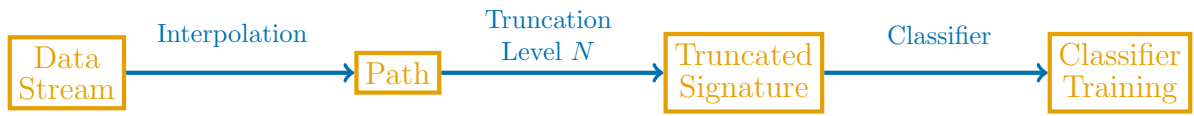
\begin{figure}[htb]\centering
	\begin{tikzpicture}[scale=2]
		\node[draw, OIorange, ultra thick, align=center](data) at (0,0) {\large Data\\ \large Stream};
		\node[draw, OIorange, ultra thick, align=center](path) at (2.2,0) {\large Path};
		\node[draw, OIorange, ultra thick, align=center](sig) at (4.5,0) {\large Truncated\\ \large Signature};
		\node[draw, OIorange, ultra thick, align=center](class) at (7,0) {\large Classifier\\ \large Training};
		\draw[OIblue, ultra thick, ->] (data) -- (path);
		\draw[OIblue, ultra thick, ->] (path) -- (sig);
		\draw[OIblue, ultra thick, ->] (sig) -- (class);
		\node[OIblue, align=center](path) at (1.1,0.25) {Interpolation};
		\node[OIblue, align=center](sig) at (3.2,0.25) {Truncation\\ Level $N$};
		\node[OIblue, align=center](class) at (5.8,0.25) {Classifier};
		\end{tikzpicture}
	\caption{The production pipeline of classification using signatures. In each step a choice has to be made: what method to use to construct the path, at which level to truncate the signature, and which classification algorithm to use.}\label{fig:exp}
\end{figure}

We are now able to put all the pieces together and build our classification algorithm. In a first step we have to take the data stream and construct a continuous path, e.g., using one of the methods discussed in Section \ref{sec:data}. Then one has to decide up to which point one wants to calculate the signature, i.e., deciding on a truncation level $N$. In this way we generate our feature set which can then be used by the classification algorithm of our choice (in \cite{IKMSS25} in particular random forests have proved successful). The full basic production pipeline can be seen in Figure \ref{fig:exp}.

Where to take it from here? For practical implementation, a good source on classification algorithms can be found in \textit{Introduction to Statistical Learning} by James, Witten, Hastie and Tibsherani \cite{JWHT23}. Algorithms that directly construct truncated signatures from data streams are \textit{{ES}ig} from \url{https://coropa.sourceforge.io/} and \textit{iisignature} \url{https://github.com/bottler/iisignature}, cf. \cite{RG18}. 

To advance the understanding of signatures, the \textit{Primer} \cite{CK16} by Chevyrev and Kormilitzin is an excellent source that goes much deeper than this introduction while being very accessible. Also the article series \cite{PNY} by P\'{e}rez on applications in Quantitative Finance is very useful. For more ideas on how to enhance the construction of path signatures from data stream, the paper \cite{MFKL21} by Morill, Fermanian, Kidger and Lyons is very helpful. 

\paragraph{Acknowledgements}
	This article is based on an introductory talk that the author gave in June 2024 for the Research Experience for Undergraduates in Industrial Mathematics and Statistics at Worcester Polytechnic Institute (WPI), funded by the National Science Foundation Award DMS-2244306. The students in this project, Yasutora Ito (WPI), Adam Mullaney (WPI) and Kathleen Shiffer (Swarthmore College) investigated the use of signatures to classify commodity futures time series by storability of the underlying commodity, \cite{IMS24}, results published in \cite{IKMSS25}. It was a pleasure and honor to mentor them and see their fast progress and growth. I am also indebted to Hari P. Krishnan who was not only the industrial sponsor for this project, but for his willingness to test new mathematical methods and eager curiosity on how mathematical models apply to practical finance problems and how these problems can in turn provide a new perspective on the mathematical methods. Thanks to Qingshuo Song and Yasutora Ito for feedback on an earlier draft.

\bibliographystyle{alpha}
\bibliography{sig-bib}

\newcommand{\etalchar}[1]{$^{#1}$}
\begin{thebibliography}{MKNH{\etalchar{+}}19}

\bibitem[Che54]{Che54}
Kuo-Tsai Chen.
\newblock Iterated integrals and exponential homomorphisms.
\newblock {\em Proc. London Math. Soc. (3)}, 4:502--512, 1954.

\bibitem[Che57]{Che57}
Kuo-Tsai Chen.
\newblock Integration of paths, geometric invariants and a generalized
  {B}aker-{H}ausdorff formula.
\newblock {\em Ann. of Math. (2)}, 65:163--178, 1957.

\bibitem[CK16]{CK16}
Ilya Chevyrev and Andrey Kormilitzin.
\newblock A primer on the signature method in machine learning, 2016.
\newblock \url{https://arxiv.org/abs/1603.03788}.

\bibitem[GLKF14]{GLKF14}
Lajos~Gergely Gyurkó, Terry Lyons, Mark Kontkowski, and Jonathan Field.
\newblock Extracting information from the signature of a financial data stream,
  2014.
\newblock https://arxiv.org/abs/1307.7244.

\bibitem[Gra13]{Gra13}
Benjamin Graham.
\newblock Sparse arrays of signatures for online character recognition, 2013.
\newblock https://arxiv.org/abs/1308.0371.

\bibitem[HL10]{HL10}
Ben Hambly and Terry Lyons.
\newblock Uniqueness for the signature of a path of bounded variation and the
  reduced path group.
\newblock {\em Ann. of Math. (2)}, 171(1):109--167, 2010.

\bibitem[IKM{\etalchar{+}}25]{IKMSS25}
Yasutora Ito, Hari~P. Krishnan, Adam Mullaney, Kathleen Shiffer, and Stephan
  Sturm.
\newblock Understanding lost civilizations and commodity term structures: The
  signature method.
\newblock {\em submitted}, 2025.

\bibitem[IMS24]{IMS24}
Yasutora Ito, Adam Mullaney, and Kathleen Shiffer.
\newblock Classifying commodities using the signature method.
\newblock Technical report, Worcester Polytechnic Institute, 2024.

\bibitem[JWH{\etalchar{+}}23]{JWHT23}
Gareth James, Daniela Witten, Trevor Hastie, Robert Tibshirani, and Jonathan
  Taylor.
\newblock {\em An introduction to statistical learning---with applications in
  {P}ython}.
\newblock Springer Texts in Statistics. Springer, Cham, 2023.

\bibitem[Lyo94]{L94}
Terry Lyons.
\newblock Differential equations driven by rough signals. {I}. {A}n extension
  of an inequality of {L}. {C}. {Y}oung.
\newblock {\em Math. Res. Lett.}, 1(4):451--464, 1994.

\bibitem[Lyo98]{L98}
Terry~J. Lyons.
\newblock Differential equations driven by rough signals.
\newblock {\em Rev. Mat. Iberoamericana}, 14(2):215--310, 1998.

\bibitem[MFKL21]{MFKL21}
James Morrill, Adeline Fermanian, Patrick Kidger, and Terry Lyons.
\newblock A generalised signature method for multivariate time series feature
  extraction, 2021.
\newblock \url{https://arxiv.org/abs/2006.00873}.

\bibitem[MKNH{\etalchar{+}}19]{MKNSHL19}
James Morrill, Andrey Kormilitzin, Alejo Nevado-Holgado, Sumanth Swaminathan,
  Sam Howison, and Terry Lyons.
\newblock The signature-based model for early detection of sepsis from
  electronic health records in the intensive care unit.
\newblock In {\em 2019 Computing in Cardiology (CinC)}, 2019.

\bibitem[P{\'{e}}rND]{PNY}
Imanol P{\'{e}}rez.
\newblock Rough path theory and signatures applied to quantitative finance.
  {P}art {I--IV}, N.D.
\newblock
  \url{https://www.quantstart.com/articles/rough-path-theory-and-signatures-applied-to-quantitative-finance-part-1/}
  [Retrieved: 7. Nov. 2024].

\bibitem[RG18]{RG18}
Jeremy Reizenstein and Benjamin Graham.
\newblock The iisignature library: efficient calculation of iterated-integral
  signatures and log signatures, 2018.
\newblock \url{https://arxiv.org/abs/1802.08252}.

\end{thebibliography}

\appendix

\section{Concatenation of Signatures and Chen's Identity}\label{sec:Chen}

In classical Riemann integration we know that we can combine intervals of integration. For a continuous function $f$ on an interval $[a,c]$, we have for an arbitrary intermediary point $b \in (a,c)$ that
\[
\int_a^b f(x) \, dx + \int_b^c f(x) \, dx = \int_a^c f(x) \, dx.
\]
The same is true for Riemann--Stieltjes integrals, for any BV function $g$ we have
\[
\int_a^b f(x) \, dg(x) + \int_b^c f(x) \, dg(x) = \int_a^c f(x) \, dg(x),
\]
which can be seen directly from the Riemann--Stieltjes sum approximation. We can thus reasonably ask that if we have a path $X$ on an interval $[a,c]$, if then also its signature is determined by the signature on the subintervals $[a,b]$ and $[b,c]$ for some intermediary point, and if so we can calculate it explicitly by adding up the signatures calculated on the subintervals. This would give us a way to concatenate signatures and calculate them stepwise over small time intervals. The answer is that we can indeed recover the signature on the entire inteval $[a,c]$ from the signatures on the subintervals, but the calculation is less obvious than in the above cases.

To see why, we might first explicitly calculate how things look at the second order level. Consider $X \, : \, [a,c] \to \R^d$ a $d$-dimensional path and $b \in (a,c)$. By the definition of the iterated integrals and the decomposition of Riemann--Stieljes integrals mentioned above we have
\begin{align*}
S^{i_1,i_2}_{a,c}(X) & = \int_a^c \int_a^{t_2} 1 \, dX^{i_1}_{t_1} dX^{i_2}_{i_2} = \int_a^b \int_a^{t_2} 1 \, dX^{i_1}_{t_1} dX^{i_2}_{i_2} + \int_b^c \int_a^{t_2} 1 \, dX^{i_1}_{t_1} dX^{i_2}_{i_2} \\ &= \int_a^b \int_a^{t_2} 1 \, dX^{i_1}_{t_1} dX^{i_2}_{i_2} + \int_b^c \biggl(\int_a^b 1 \, dX^{i_1}_{t_1} + \int_b^{t_2} 1 \, dX^{i_1}_{t_1} \biggr)dX^{i_2}_{i_2} \\ &= \int_a^b \int_a^{t_2} 1 \, dX^{i_1}_{t_1} dX^{i_2}_{i_2} + \int_b^c \int_a^b 1 \, dX^{i_1}_{t_1} dX^{i_2}_{i_2} + \int_b^c \int_b^{t_2} 1 \, dX^{i_1}_{t_1} dX^{i_2}_{i_2} \\&=. 
\int_a^b \int_a^{t_2} 1 \, dX^{i_1}_{t_1} dX^{i_2}_{i_2} + \int_a^b 1 \, dX^{i_1}_{t_1} \cdot \int_b^c  1 \,  dX^{i_2}_{i_2} + \int_b^c \int_b^{t_2} 1 \, dX^{i_1}_{t_1} dX^{i_2}_{i_2} \\
& = S^{i_1,i_2}_{a,b}(X) + S^{i_1}_{a,b}(X)S^{i_2}_{b,c}(X) + S^{i_1,i_2}_{b,c}(X).
\end{align*}
We note that this formula is not only relying on the second order signature terms from the subintervals, but also on the  first order terms. Moreover, it is not symmetric as only $S^{i_1}_{a,b}(X)S^{i_2}_{b,c}(X)$ appears and not $S^{i_2}_{a,b}(X)S^{i_1}_{b,c}(X)$. This is due to the fact that the order of the integration matters. We can develop higher order terms recursively, e.g.,
\begin{align*}
S^{i_1,i_2, i_3}_{a,c}(X) & = \int_a^c S^{i_1,i_2}_{a,t_3}(X) \, dX^{i_3}_{i_t} = \int_a^b S^{i_1,i_2}_{a,t_3}(X) \, dX^{i_3}_{i_t} + \int_b^c S^{i_1,i_2}_{a,t_3}(X) \,  dX^{i_3}_{i_t}\\& =  \int_a^b S^{i_1,i_2}_{a,t_3}(X) \, dX^{i_3}_{i_t} + \int_b^c S^{i_1,i_2}_{a,b}(X) \, dX^{i_3}_{i_t}  + \int_b^c  S^{i_1}_{a,b}(X) S^{i_2}_{b,t_3}(X) \,  dX^{i_3}_{i_t} + \int_b^c S^{i_1,i_2}_{b,t_3}(X)  \,dX^{i_3}_{i_t} \\ &= S^{i_1,i_2, i_3}_{a,b}(X) + S^{i_1,i_2}_{a,b}(X)S^{i_3}_{b,c}(X) + S^{i_1}_{a,b}(X)S^{i_2,i_3}_{b,c}(X) + S^{i_1,i_2, i_3}_{b,c}(X).
\end{align*}
 Based on this, we can conjecture the general form, which then can be proved by induction:
\begin{equation}\label{eq:Chen}
S^{i_1,i_2,\ldots, i_{n-1}, i_n}_{a,c}(X) = \sum_{k=0}^n S^{i_1,i_2,\ldots, i_{n-1}, i_k}_{a,b}(X) S^{i_{k+1}1,i_2,\ldots, i_{n-1}, i_n}_{b,c}(X).
\end{equation}
This formula is known as \textit{Chen's identity}. Note that here the zeroth order terms ($k=0$ in the first factor and $k=n$ in the second) are defined to equal $1$, the zeroth order term of the signature.

For truncated signatures we have a slightly different picture. While knowledge about the level-$N$ truncated signatures on the subintervals $[a,b]$ and $[b,c]$ allows us to calculate the level-$N$ truncated signature on $[a,c]$ by Chen's identity, they contain actually more information. Again, from Chen's identity we see that they also contribute to higher order terms of the signature on the full interval. While this information is not enough to reconstruct these higher order signature terms completely, it provides partial information, e.g., $2N$ out of the $2N+2$ subinterval terms that make up the level $N+1$-signature are of level $N$ or less.

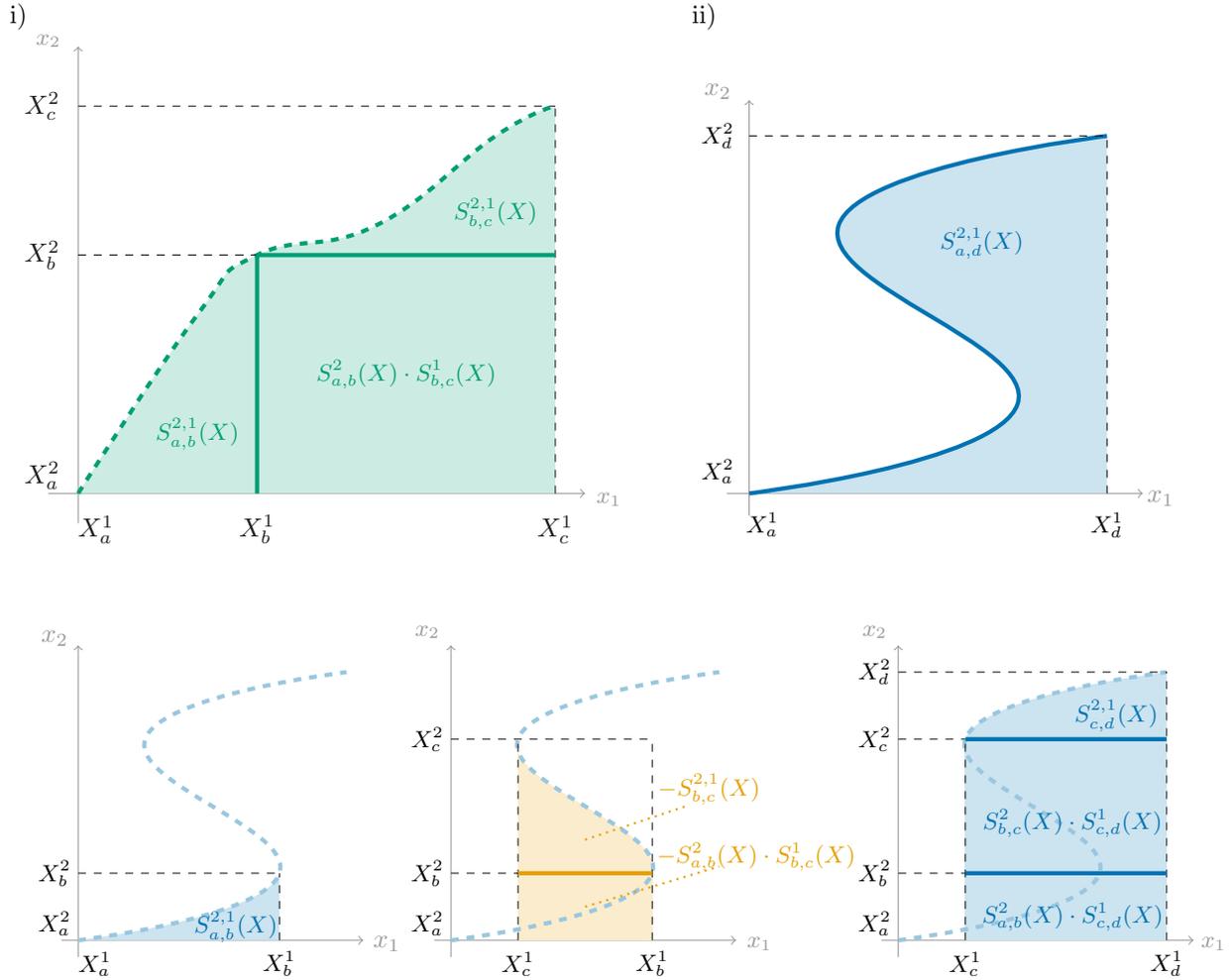
\begin{figure}[htb]
\centering
	\begin{tikzpicture}
		\begin{scope}[scale=0.8]
		\node at (-1cm,8cm) {i)};
		\draw[gray!80, ->] (-0.5cm,0cm) -- (8.5cm,0cm) node at (8.9, -0.1) {\small $x_1$};  
		\draw[gray!80, ->] (0cm,-0.5cm) -- (0cm,7.5cm) node at (-0.5, 7.6) {\small $x_2$};  
		\fill[OIgreen!20] (0,0) -- plot[smooth, tension=0.8] coordinates {(0,0) (2,3) (3,4) (5,4.5) (7,6) (8,6.5)} -- (8,0) -- cycle;
		\draw[OIgreen, ultra thick, dashed] plot[smooth, tension=0.8] coordinates {(0,0) (2,3) (3,4) (5,4.5) (7,6) (8,6.5)};
		\draw[black!90, dashed] (0,4) -- (3,4) (3,0) -- (3,4) (8,0) -- (8,6.5) (0,4) -- (8,4)  (0,6.5) -- (8,6.5)node at (-0.6,0.3) {$X^2_a$} node at (-0.6,4) {$X^2_b$} node at (-0.6,6.5) {$X^2_c$} node at (0.3,-0.6) {$X^1_a$} node at (3,-0.6) {$X^1_b$} node at (8,-0.6) {$X^1_c$};
		\draw[OIgreen, ultra thick] (3,0) -- (3,4) (3,4) -- (8,4) node[OIgreen] at (2,1) {\small $S^{2,1}_{a,b}(X)$} node[OIgreen] at (7,4.7) {\small $S^{2,1}_{b,c}(X)$} node[OIgreen] at (5.5,2) {\small $S^{2}_{a,b}(X) \cdot S^{1}_{b,c}(X)$};
		\end{scope}
		\begin{scope}[xshift=9cm, scale=0.6]
		\node at (-1cm,10.67cm) {ii)};
		\draw[gray!80, ->] (-0.5cm,0cm) -- (8.8cm,0cm) node at (9.2, -0.1) {$x_1$};  
		\draw[gray!80, ->] (0cm,-0.5cm) -- (0cm,8.8cm) node at (-0.7, 9) {$x_2$};  
		\fill[OIblue!20] (0,0) -- plot[smooth, tension=0.8] coordinates {(0,0) (6,2) (2,6) (8,8)} -- (8,0) -- cycle;
		\draw[OIblue, ultra thick] plot[smooth, tension=0.8] coordinates {(0,0) (6,2) (2,6) (8,8)};
		\node at (-0.7,0.5) {\small $X^2_a$} node at (0.3,-0.7) {\small $X^1_a$} node at (8,-0.7) {\small $X^1_d$} node at (-0.7,8) {\small $X^2_d$};
		\draw[black!90, dashed] (0,8) -- (8,8) (8,0) -- (8,8)  node[OIblue] at (5.2,5.6) {\small $S^{2,1}_{a,d}(X)$} ;
		\end{scope}
		\begin{scope}[xshift=0cm, yshift=-6cm, scale=0.45]
		\draw[gray!80, ->] (-0.5cm,0cm) -- (8.5cm,0cm) node at (9.2, -0.1) {$x_1$};  
		\draw[gray!80, ->] (0cm,-0.5cm) -- (0cm,8.8cm) node at (-0.7, 9) {$x_2$};  
		\begin{scope}
		\clip (0,0) -- plot[smooth, tension=0.8] coordinates {((0,0) (6,2) (2,6) (8,8)} -- (8,0) -- cycle;;
		\fill[OIblue!20] (0,0) -- plot[smooth, tension=0.8] coordinates {(0,0) (6,2)} -- (6,0) -- cycle;
		\end{scope}
		\draw[OIblue!40, ultra thick, dashed] plot[smooth, tension=0.8] coordinates {(0,0) (6,2) (2,6) (8,8)};
		\node at (-0.7,0.5) {\small $X^2_a$} node at (0.5,-0.7) {\small $X^1_a$} node at (6,-0.7) {\small $X^1_b$} node at (-0.7,2) {\small $X^2_b$};
		\draw[black!90, dashed] (0,2) -- (6,2) (6,0) -- (6,2)  node[OIblue] at (4.7,0.4) {\small $S^{2,1}_{a,b}(X)$} ;
		\end{scope}
		\begin{scope}[xshift=5cm, yshift=-6cm, scale=0.45]
		\draw[gray!80, ->] (-0.5cm,0cm) -- (8.5cm,0cm) node at (9.2, -0.1) {\small $x_1$};  
		\draw[gray!80, ->] (0cm,-0.5cm) -- (0cm,8.8cm) node at (-0.7, 9.2) {\small $x_2$};  
		\begin{scope}
		\fill[OIorange!20] (2,0) -- (6,0) -- (6,2) -- (2,2) -- cycle;
		\clip (0,0) -- plot[smooth, tension=0.8] coordinates {((0,0) (6,2) (2,6) (8,8)} -- (0,8) -- cycle;;
		\fill[OIorange!20] (2,0) -- (6,0) -- (6,6) -- (2,6) -- cycle;
		\end{scope}
		\draw[OIblue!40, ultra thick, dashed] plot[smooth, tension=0.8] coordinates {(0,0) (6,2) (2,6) (8,8)};
		\node at (-0.7,0.5) {\small $X^2_a$}  node at (-0.7,6) {\small $X^2_c$} node at (2,-0.7) {\small $X^1_c$} node at (6,-0.7) {\small $X^1_b$} node at (-0.7,2) {\small $X^2_b$};
		\draw[black!90, dashed] (0,6) -- (6,6) (6,0) -- (6,6) (2,0) -- (2,6) (0,2) -- (6,2) node[OIorange] at (7.7,4.5) {\small$-S^{2,1}_{b,c}(X)$} node[OIorange] at (9.1,2.5) {\small $-S^{2}_{a,b}(X) \cdot S^{1}_{b,c}(X)$};
		\draw[OIorange, ultra thick] (2,2) -- (6,2);
		\draw[OIorange, thick, dotted] (4,3) -- (7,4) (4,1) -- (8,2.2);
		\end{scope}
		\begin{scope}[xshift=11cm, yshift=-6cm, scale=0.45]
		\draw[gray!80, ->] (-0.5cm,0cm) -- (8.8cm,0cm) node at (9.4, -0.1) {\small $x_1$};  
		\draw[gray!80, ->] (0cm,-0.5cm) -- (0cm,8.8cm) node at (-0.7, 9.2) {\small $x_2$};  
		\begin{scope}
		\clip (0,0) -- plot[smooth, tension=0.8] coordinates {((0,0) (6,2) (2,6) (8,8)} -- (8,0) -- cycle;;
		\fill[OIblue!20] (2,0) -- (8,0) -- (8,8) -- (2,8) -- cycle;
		\end{scope}
		\fill[OIblue!20] (2,0) -- (8,0) -- (8,6) -- (2,6) -- cycle;
		\draw[OIblue!40, ultra thick, dashed] plot[smooth, tension=0.8] coordinates {(0,0) (6,2) (2,6) (8,8)};
		\node at (-0.7,0.5) {\small $X^2_a$}  node at (-0.7,2) {\small  $X^2_b$} node at (-0.7,6) {\small  $X^2_c$} node at (2,-0.7) {\small  $X^1_c$}  node at (8,-0.7) {\small  $X^1_d$} node at (-0.7,8) {\small  $X^2_d$};
		\draw[black!90, dashed] (0,8) -- (8,8) (8,0) -- (8,8) (0,6) -- (2,6) (2,0) -- (2,6) (0,2) -- (6,2) node[OIblue] at (6.5,6.7) {\small$S^{2,1}_{c,d}(X)$} node[OIblue] at (5.2,3.5) {\small $S^{2}_{b,c}(X) \cdot S^{1}_{c,d}(X)$} node[OIblue] at (5.2,0.7) {\small $S^{2}_{a,b}(X) \cdot S^{1}_{c,d}(X)$};
		\draw[OIblue, ultra thick] (2,2) -- (8,2) (2,6) -- (8,6);
		\end{scope}
	\end{tikzpicture}
	\caption{Geometric interpretation of Chen's identity. i) If the path is increasing in each component, areas just add up: $S^{2,1}_{a,c}(X) = S^{2,1}_{a,b}(X) + S^{2}_{a,b}(X)S^{1}_{b,c}(X) + S^{2,1}_{b,c}(X)$. ii) In general we have to add and subtract the areas, as the terms in Chen's formula have different signs: the area in the upper right graph is composed by the signed areas of the graphs below.}\label{fig:Chen}
\end{figure}

Chen's identity can help us also to understand the iterated integrals appearing in the signature geometrically. Remember we said that in the case that the path is increasing in each dimension, the signature corresponds to the area under the path (in two dimensions) or the volume spanned by the path (in higher dimensions). If the path is monotone (but not necessarily increasing), this affects the sign of the area: if the number of decreasing components is even, it is positive, if it is odd, the sign is negative. For the general case we can use Chen's formula to split the path into subintervals that we concatenate. In the nice case of a component-wise increasing paths, this means just splitting up the area (or volume), see Figure \ref{fig:Chen} a). In the general case, we split the path into subintervals on which it is monotone, see, e.g., Figure \ref{fig:Chen} b). Applying Chen's formula repeatedly we have
\begin{align*}
S^{2,1}_{a,d}(X) & = S^{2,1}_{a,c}(X) + S^{2}_{a,c}(X)S^{1}_{c,d}(X) + S^{2,1}_{c,d}(X)\\
& = \Bigl(S^{2,1}_{a,b}(X) + S^{2}_{a,b}(X)S^{1}_{b,c}(X) + S^{2,1}_{b,c}(X)\Bigr)  + \Bigl(S^{2}_{a,b}(X)+ S^{2}_{b,c}(X)\Bigr) S^{1}_{c,d}(X) + S^{2,1}_{c,d}(X)\\
& = S^{2,1}_{a,b}(X) + \Bigl(S^{2}_{a,b}(X)S^{1}_{b,c}(X) + S^{2,1}_{b,c}(X)\Bigr)  + \Bigl(S^{2}_{a,b}(X)S^{1}_{c,d}(X)+ S^{2}_{b,c}(X)S^{1}_{c,d}(X)  + S^{2,1}_{c,d}(X)\Bigr)
\end{align*}
Here the path is in the intervals $[a,b]$ and $[c,d]$ increasing in all components, so all terms here have positive sign. On the interval $[b,c]$, however, the path is decreasing in the first component and increasing in the second, so the associated lengths and areas are negative. In the lower row of Figure \ref{fig:Chen} we see how the signed areas add up to the total area.

\end{document}